\documentclass[11pt,a4paper,english]{article}
\usepackage[utf8]{inputenc} 

\usepackage{babel}

\usepackage{stmaryrd}

\usepackage[noblocks]{authblk}

\usepackage{geometry}
\usepackage{amsmath, amssymb}

\usepackage[autostyle, english = american]{csquotes}
\MakeOuterQuote{"}

\usepackage{subcaption}
\usepackage{makecell}       
\usepackage{booktabs}       
\usepackage{array}          
\usepackage{tabularx} 
\usepackage{tabularx, makecell} 
\usepackage[table]{xcolor} 
\definecolor{rowgray}{gray}{0.96} 
\renewcommand{\arraystretch}{1.4}
\newcolumntype{Y}{>{\centering\arraybackslash}X}
\usepackage{caption}
\usepackage[font=footnotesize]{caption}
\usepackage{float}
\usepackage{graphicx}
\usepackage{multirow}
\usepackage{caption} 
\usepackage{tikz-cd}
\usetikzlibrary{arrows.meta,positioning,fit,backgrounds}
\usepackage{xcolor}
\definecolor{CHCtitle}{HTML}{0A3D62}   
\definecolor{CHCbox}{HTML}{EEF5FF}     
\definecolor{CHCline}{HTML}{2C3E50}    
\definecolor{CHCalt}{HTML}{FFF7E6}     

\usepackage{natbib}

\usepackage{caption}
\setcitestyle{aysep={}} 
\title{\textbf{Structural Ranking of the Cognitive Plausibility of Computational Models of Analogy and Metaphors with the Minimal Cognitive Grid}}
\author[1]{Alessio Donvito}
\author[2]{Antonio Lieto}
\affil[1]{University of Bari, Italy}
\affil[2]{University of Salerno, CIIT Lab / ICAR-CNR, Italy}

\date{To appear in Cognitive Systems Research}

\begin{document} 
\maketitle

\begin{abstract}
 In this paper, we employ the Minimal Cognitive Grid (MCG), a framework created to evaluate the cognitive plausibility of artificial systems, to offer a systematic assessment of leading computational models of analogy and metaphor, including the Structure-Mapping Engine (SME), CogSketch, MET\textsuperscript{CL}, and Large Language Models (LLMs). We present a formal and quantitative operationalization of the MCG framework and, through the analysis of its three main dimensions (Functional/Structural Ratio, Generality, and Performance Match), examine how well each system aligns with standard cognitive theories of the modeled phenomena, thus allowing for comparison of the models with respect to their cognitive plausibility, according to consistent and generalizable mathematical criteria.
\end{abstract}

\section*{Introduction}

The creation, use, and interpretation of analogies and metaphors expressed in natural language
sentences represent a crucial capacity of human semantic competence and communication. Analogical reasoning is widely regarded as a foundational cognitive process that underlies our ability to learn, generalize, and solve problems. By recognizing correspondences between different concepts and domains, individuals can extend existing knowledge to unfamiliar situations, identify regularities across varied contexts, and provide innovative solutions to new challenges. Given its relevance, uncovering the computational basis of analogy has been a long-standing objective in cognitive sciences. In the context of computational cognitive modeling, many different approaches have been proposed to endow artificial systems with the same ability. In this paper, we first review the main, state-of-the-art, computational models of analogy and metaphor through the epistemological lenses of computational cognitive science. In addition,
we analyze and compare the following AI systems: 1) the Structure-Mapping Engine (SME) [\cite{falkenhainer1989structure}]: this is one of the most influential systems in the field of analogical reasoning, whose underlying principles of identifying systematic correspondences between different domains have also been extended to metaphor understanding (where metaphors are seen as a form of analogy);  2) CogSketch [\cite{cogsketch}]: an AI system that provides a sketch understanding environment that supports visual analogical reasoning and enables users to draw diagrams that it interprets, supporting reasoning about both physical and abstract concepts; 3)MET\textsuperscript{CL} [\cite{lieto2025delta}]: a compositional logic-based AI model for metaphor generation; 4) Large Language Models (in particular GPT-3.5, GPT-4, and GPT-4o) used for analogy and metaphor comprehension and generation (currently, no systematic assessment on such tasks of more recent systems like GPT 5.2 is available in the literature). In this paper we evaluate each of these systems with the Minimal Cognitive Grid: a pragmatic framework proposed by [\cite{lieto2021cognitive}] to rank the different degrees of structural accuracy of artificial systems, in order to predict their explanatory power. In the section below, we introduce the theoretical distinction between functional and structural artificial models, along with the main features of the Minimal Cognitive Grid (MCG). Then, we quantitatively operationalize its dimensions and analyze each system on them. Finally, we combine each analysis within an overall cognitive plausibility score, showing that the MCG offers a blueprint for a generalizable comparison and ranking of artificial systems and models according to their biological or cognitive plausibility. The paper concludes with remarks and directions for future work.

\section{Functional/Structural Models and the Minimal Cognitive Grid}

An important distinction in AI and computational cognitive modeling can be established between functional and structural models (or better: between functionally and structurally designed models). 
Functionalism was introduced in Philosophy of Mind by Hillary Putnam in his seminal article titled Minds and Machines [\cite{putnam1960minds}]. This approach had a direct influence on AI, since it led to the definition of a design approach based on the notion of “functional equivalence” between some cognitive faculties in humans and the corresponding mechanisms implemented in AI programs. In other words, it posited that, from an explanatory point of view, the relation between “natural mind” and “artificial software” could have been based purely on a macroscopic equivalence of the functional organization of the two systems and their input-output specification. This position has been widely criticized in the literature over the last decades. In particular, models and systems designed according to the “functionalist” perspective are not good candidates for advancing the science of cognitive AI since, as in the case of an airplane, which flies through mechanisms entirely different from those of a bird [\cite{russell1995artificial}], the mechanisms and the overall design choices adopted to build such artifacts prevent them from having any kind of explanatory role with respect to their counterparts in nature. This is the case, for example, of recent AI technologies, including some self-proclaimed “cognitive computing” systems like IBM Watson, AlphaGo, or Large Language Models [\cite{loru2025simulation}]. In short, such systems, as the vast majority of current AI systems, are “functional” artifacts. They “function as” a natural system (in terms of the provided output given the input that they process) but the internal mechanisms determining that output are completely different with respect to what we do as humans. Therefore, a mere artificial imitation of cognitive capabilities does not necessarily function according to the same principles. 

Differently from functionalism and functionally designed systems, the structural approach to model building in AI calls for a more strongly constrained connection between the designed artificial systems, and their internal architecture and procedures, and the corresponding ones available in a target biological system. According to such an approach, structurally constrained artificial models and systems can be useful both to advance the science of AI in terms of technological achievements (e.g., in tasks that are easily solvable for humans but very hard to solve for machines using non-cognitive inspired approaches, such as, for example, in commonsense reasoning) and to play the role of “computational experiments”, able to provide insights and results useful in refining or rethinking theoretical aspects concerning the target natural system used as a source of inspiration [\cite{cordeschi2002discovery}, \cite{milkowski2013explaining}].
An immediate problem arising in this view is that it is not possible to build a completely structural and constrained artificial model, since it is not possible to reproduce a realistic artificial “replica” of a natural system. The search for increasingly more structural models, in fact, leads to an asymptotic regress to the microscopic physical world, until it reaches the well-known Wiener paradox, summarized in the remark that “The best material model of a cat is another, or preferably the same, cat”  [\cite{rosenblueth1945role}]. In short, this “paradox” points to the need for constructing proxy models, not replicas, of a given natural system, while emphasizing the difficulty of such a challenge. 
A way to address this state of affairs consists in considering the functionalism/structuralism dichotomy as the two extremes of a continuum. Between the explanatory limitations of purely functional artificial models (that, however, can achieve impressive performances in specific tasks, like shown by modern AI systems), and the impracticability of pure structural models, it is possible to identify a wide range of plausible artificial proxy models having different degrees of explanatory power with respect to the natural systems that they take as sources of inspiration \footnote{The functional/structural continuum applies both to symbolic and sub-symbolic systems, since it concerns the design choices and the biological/cognitive constraints embedded within each of these modeling frameworks. For example, in the field of Artificial Neural Networks, some architectures move along the functional axis (these include current deep learning architectures, which are considered biologically implausible for a number of reasons, such as the use of gradient-based learning methods (e.g., backpropagation), which have long been argued to lack biological plausibility [\cite{crick1989recent}]), while others include bio-inspired neural models. Among the most interesting developments in recent years, in this respect, are spiking neural models, Hierarchical Temporal Memory (HTM), and so-called evo-devo networks, which incorporate evolutionary computing algorithms that allow for the adaptation of the system to changing environments and provide greater robustness and optimization capabilities for specific classes of problems. As in the case of connectionist modeling frameworks, also symbolic systems can be analyzed through the functional/structural continuum.
In this respect, probabilistic, fuzzy, and non-monotonic reasoning formalisms increase the functional similarity between human and artificial cognitive capabilities, since they are more flexible than classical logic, but there is no evidence that they are structurally adequate models of human reasoning and understanding; in many cases, there is even direct or indirect evidence to the contrary.}. 

Within this context, the Minimal Cognitive Grid (MCG) \cite{lieto2021cognitive} was introduced to quantitatively and qualitatively assess the level of structural accuracy (i.e. the level of biological or cognitive plausibility) of artificial systems. In a nutshell, the MCG considers three key dimensions that characterize the relationship between a model and its biological or cognitive target system: 1. Functional/Structural Ratio: it concerns the balance between functional and structural components in the model, evaluating the extent to which it relies on mere functional replication versus structural mechanisms; 2. Generality: this dimension assesses the breadth of phenomena that the model can represent. A highly general model can be applied to a wide range of cognitive functions or biological systems, while a narrow model is tailored to a specific task or domain; 3. Performance Match: this dimension involves a direct comparison between the model's performance and that of the target system. It considers not only the overall accuracy of the model but also the similarity of its errors and execution times to those of the biological or cognitive system. A close performance match suggests a higher degree of psychological or biological plausibility.  In the following sections, each of the 
dimensions characterizing the MCG is operationalized, and the different computational models of analogical and metaphor processing are ranked accordingly.

\section{Functional/Structural Ratio}

In this section, we define and apply a metric ($FSR_\mathcal{M}$) to measure the ratio between functionality and structurality in the analyzed computational models of analogy and metaphor. $FSR_\mathcal{M}$ allows us to evaluate how a given computational model $\mathcal{M}$ adheres to one or more selected cognitive theories of analogical and metaphorical reasoning \{$\mathcal{T}_1, \dots, \mathcal{T}_n $\}, instead of merely "functioning" on analogical tasks. In other terms, our goal is to measure the extent to which the models' performances are achieved through a computational implementation accurate to the cognitive mechanisms postulated by the underlying theory, with distinct theories offering diverging accounts of how information is represented and processed in the human mind.

For the cognitive phenomenon of analogy, the Structure Mapping Theory [\cite{gentnerstructuremapping}] is often used as a standard.
SMT describes how the characteristics of a familiar object or process (base domain) are used for understanding a new and unknown idea (target domain). For instance, the flow of electrons in a circuit (target domain) is analogous to the flow of people in subway tunnels (base domain). If a resistor is added to the circuit, the flow of electrons decreases in intensity, in the same way that if a tight barrier is installed at the entrance of the subway, the flow of people decreases accordingly. In the example, the elements of the base and target domains are hierarchically organized in a causal structure. Between them, the analogy establishes a two-way correspondence (person-electron, barrier-resistor, people movement-electric flow), although the elements are entirely different. Unlike other theories of analogy [\cite{tversky1977features}], the SMT does not involve carrying out relational mapping between \emph{objects} in the base and target domains; instead, the mapping involves the \emph{relationships (or structures) between objects} within the hierarchy of the two sets. Gentner's Structure Mapping approach has also been extended to metaphor as a special case of analogy [\cite{gentner1988metaphor}, \cite{gentner2008metaphor}].

In contrast, the Categorization Theory of Metaphor [\cite{glucksberg2001understanding}; \cite{hampton1987inheritance}] conceptualizes metaphorical meaning as the result of class-inclusion processes grounded in property inheritance. The core assumption here is that metaphorical statements of the form “X is Y” are interpreted not as elliptical comparisons (i.e., X is like Y), but rather as assertions of category membership. For instance, the metaphorical sentence "My job is a jail" implies that “my job” is a member of a conceptually constructed “jail” category, inheriting properties such as "restriction" or "lack of freedom", rather than suggesting a literal similarity.

Each of these theoretical frameworks is supported by different corpora of psychological and linguistic evidence [\cite{holyoak2018metaphor}], and has informed the architecture of computational models that embody their principles in operational form. Given this theoretical pluralism, it would be epistemologically inconsistent to evaluate the structural plausibility of the analyzed computational systems based exclusively on the constraints of one cognitive theory. This methodological requirement involves two corollaries. First, multiple theories must be analyzed to extract cognitively justified structural constraints. Second, all models under evaluation must be tested against all the identified constraints, allowing for both intra-theoretical and cross-theoretical comparisons. In this framework, we define structural constraints both for SMT-based models and for systems grounded in the Categorization Theory of Metaphor.

Starting from the SMT framework, we identify four key principles that a model should implement to be considered structurally plausible: \textbf{1) One-to-one mapping}: each element in the \emph{source} domain can be mapped to \emph{at most one element} in the \emph{target} domain, and vice versa:

\[
\forall x \in \text{Source},\ \exists!\, y \in \text{Target} \ \text{such that}\ \text{Map}(x) = y
\]

\noindent
The one-to-one mapping restriction has been introduced to avoid ambiguity and overlap during the analogy process [\cite{gentnermarkman}]. A computational model implements this principle if it imposes the one-to-one constraint between source and target units and rejects or penalizes multiple correspondences; \textbf{2) Parallel connectivity}: if two predicates (relations) are mapped to each other, then their respective arguments must also be mapped in parallel:

\[
\text{Map}(R(x_1, x_2)) = R'(y_1, y_2) \Rightarrow \text{Map}(x_1) = y_1 \land \text{Map}(x_2) = y_2
\]

\noindent
The principle is verified if the model rejects correspondences between relations when the arguments are not also mappable in parallel, and it can be understood as a structural extension of the more fundamental one-to-one mapping constraint, which ensures that each element in the source maps to exactly one element in the target, but does not, on its own, guarantee the internal coherence of these correspondences. Parallel connectivity strengthens this foundation by requiring that, whenever two relations are aligned between domains, their corresponding arguments must also be systematically mapped; \textbf{3) Systematicity}: interconnected relational structures are favored over isolated relations:

\[
\text{Prefer}(M_1) > \text{Prefer}(M_2)\quad \text{if} \quad \text{Relations}_{M_1} \supset \text{Relations}_{M_2}
\] 

\noindent
This is the guide principle of SMT: the human mind favors deep and coherent mappings rather than superficial associations [\cite{gentnerstructuremapping}; \cite{gentner2001analogical}]. The principle is verified if the model generates mappings that preserve interconnected relational structures rather than simple matching between attributes [\textit{e.g.} $\textit{Taller}(Maria, Luisa)\to \textit{Taller}(Anna, Carla)$ rather than $\textit{Tall}(Maria) \to \textit{Tall}(Anna)$]; \textbf{4) Inferential projection}: the model must support novel inferences in the target domain based on the mapped structure.
Inferential projection distinguishes analogy from sheer comparison: analogy is productive, allowing to infer previously unobserved properties [\cite{gentnerstructuremapping}; \cite{falkenhainerSME}]. The principle's criterion for observability is whether the model can detect new relationships in the target that were not explicitly contained in the source.

For the Categorization Theory of Metaphor, we draw on theoretical analyses by [\cite{holyoak2018metaphor}, \cite{hampton1987inheritance}, \cite{glucksberg2001understanding}], and identify two core structural principles: \textbf{1) Categorization}. The central claim of CTM is that metaphors are categorization statements. Let us consider the classic example: "My lawyer is a shark". Here, the \emph{vehicle} (“shark”) doesn’t refer to the literal animal but to an abstract ad hoc category (for instance: “vicious, aggressive beings”). The \emph{target} ("lawyer") is interpreted as a member of this abstract class. Thus, given a metaphor of the form "$T$ is $V$", an ad hoc category $C(V)$ is evoked by V, with $T\in C(V)$. A computational model satisfies this constraint if it generates $C(V)$ and assigns $T$ to it meaningfully; \textbf{2) Property Selection}. CTM posits that metaphorical meaning arises from the selective projection of properties from the vehicle category to the target, so that only significant features are transferred: if $C(V)=\{ p_1, p_2, ..., p_n\}$ denotes the set of characteristics of the category $C(V)$, and $R\subseteq C(V)$ is a relevant subset of features selected for transfer, then the \emph{target} inherits at least the relevant characteristics of the vehicle category ($Props(T)\supseteq R$).


In order to measure the ratio between functionality and structurality in our computational models, given a set of $k$ theoretical constraints $\{c_1, \dots, c_k\}$ derived from cognitive theories $\{ \mathcal{T}_1, \dots, \mathcal{T}_n \}$, each model $\mathcal{M}$, regardless of whether it was designed with reference to $\{ \mathcal{T}_1, \dots, \mathcal{T}_n \}$, is then evaluated against the full set of constraints \( \{c_1, \dots, c_k\} \). For each constraint \( c_i \), the evaluation involves two components:

\begin{enumerate}
  \item \textbf{Partial structural score} \( s_i \in \{0, 1\} \), indicating whether the model implements the structural mechanism posited by the constraint:
  
  \begin{equation}
  s_i =
  \begin{cases}
    1 & \text{if $\mathcal{M}$ implements the structural mechanism} \\
    0 & \text{if $\mathcal{M}$ does not implement the mechanism structurally}
  \end{cases}
\end{equation}

  \item \textbf{Weight} \( w_i \in [0, 1] \), reflecting the relative importance of constraint \( c_i \) within the theory, with the normalization condition \( \sum_i w_i = 1 \).
\end{enumerate}

\vspace{1em}

\noindent
At this stage Since we consider functionality as the complement of structurality, we define the partial functional score simply as:  

\begin{equation}
f_i = 1 - s_i
\end{equation}

\noindent
Here, we aim at assessing whether the target  mechanisms are implemented or not by the evaluated systems; a binary scoring scheme captures this objective. This means that if a constraint is structurally implemented (\(s_i = 1\)), it cannot be considered functionally satisfied at the same time, and vice versa. We can then compute the overall Structural and Functional scores (${S_{M}}, {F_{M}}$) as:

\begin{equation}
  S_M = \sum_{i=1}^{k} w_i s_i, 
  \quad
  F_M = \sum_{i=1}^{k} w_i f_i = 1 - S_M
\end{equation}

\noindent
And define the $\boldsymbol{FSR_\mathcal{M}}$ as:

\begin{equation}
  {FSR_\mathcal{M}} = \frac{F_M}{S_M + \varepsilon}
  = \frac{1 - S_M}{S_M + \varepsilon}
  \hspace{1cm}
\end{equation}

\noindent
The constant $\varepsilon = 0.01$ is introduced for regularization purposes, preventing division by zero in cases where $S_M = 0$, while minimally affecting the function over most of its domain. This choice can affect $FSR_\mathcal{M}$ if $S_M$ is low; however, as discussed below, its impact becomes negligible after normalization. Also, the effect of the non-linear formulation of $FSR_\mathcal{M}$ is observable only at the level of the raw scores, where it makes the magnitude of the differences between the models more apparent. This non-linearity does not materially persist in the final normalized values anyway, since the final score reduces to $S_{\mathcal M}$, with $\varepsilon$ introducing only a small offset. Values of ${FSR_\mathcal{M}}<1$ indicate a high degree of structural plausibility, whereas values $>1$, and progressively more so as they increase, indicate that the system deviates from the internal constraints posited by $T$.

\subsection{SME/CogSketch}

\subsubsection*{One-to-one mapping}

In SME, one-to-one mapping is enforced by the \emph{Global Mapping} construction phase [\cite{falkenhainerSME}]. During the generation of match hypotheses (MHs), mappings that violate the bijection constraint are filtered out. Match sets containing multiple mappings to the same target entity are penalized and excluded during the selection of the optimal Global Mapping (\textit{gmap}). CogSketch extends SME by incorporating visual-spatial reasoning, yet retains SME as its core analogical engine. When produced in CogSketch, the mapping process operates over “glyphs” — the system’s representation of drawn objects — paired with conceptual labels from a structured ontology (typically OpenCyc). The system represents each sketch as a structured description group (Dgroup), which includes symbolic representations of the glyphs (drawn entities) and the spatial relations among them. When comparing two such Dgroups—e.g., a cup above a table in the base sketch and a bowl above a counter in the target—SME attempts to construct match hypotheses between corresponding entities and relations. These hypotheses are then assembled into candidate global mappings. As reported in [\cite{forbus2008cogsketch}; \cite{lovett2007analogy}] the one-to-one constraint is shown to govern the interpretation of visual analogies, including spatial tasks like Raven’s Progressive Matrices. In these contexts, each glyph representing an object or relational structure in the visual field is mapped to a unique counterpart in the analogized scene.

\subsubsection*{Parallel connectivity}

The principle of parallel connectivity imposes a consistency constraint on relational alignment: if a relation $R(x_1, x_2, ..., x_n) $ in the base domain is mapped to a relation $R'(y_1, y_2, ..., y_n) $ in the target domain, then for the mapping to be valid, each argument $x_i$ must correspond to its counterpart $y_i$. 

In SME, the principle is explicitly part of the  match hypotheses construction process. During the generation of candidate relational mappings, SME inspects the arguments of the mapped predicates and checks whether those arguments are themselves suitable for being mapped to each other. If such correspondence is absent or violated — \textit{i.e.}, if a pair of mapped predicates links arguments that are mismatched — then the match hypothesis is invalidated and excluded from the set of possible global mappings.
CogSketch inherits this constraint via its use of SME as its analogical reasoning module. Visual entities in CogSketch are annotated with conceptual labels and organized into symbolic relational descriptions. When such predicates (\textit e.g., LEFT-OF, ABOVE) are compared across two sketched scenes, their arguments (the glyphs representing objects) must be matched accordingly for the relation to be accepted as part of the analogical mapping. For example, the mapping of LEFT-OF(Cup, Plate) in one sketch to LEFT-OF(Bowl, Spoon) in another presupposes that $Cup \leftrightarrow Bowl$ and $ Plate \leftrightarrow Spoon $ are valid individual mappings.
In both SME and CogSketch, then, parallel connectivity is implemented and functional to the generation of valid analogical mappings: neither system allows for isolated predicate alignment without corresponding argument-level structure.

\subsubsection*{Systematicity}

SME implements systematicity in the computation of a Structural Evaluation Score (SES), which measures the quality of the global mapping produced. During the mapping process, SME assembles consistent match hypotheses into candidate global mappings (gmaps), then evaluates each gmap by summing the weights of its constituent match hypotheses. Crucially, these weights are not uniform: SME assigns greater weight to matches developed upon larger, more interconnected relational structures. In other words, mappings that participate in higher-order relations contribute more to the SES than matches of singular attributes. In the authors' example [\cite{falkenhainerSME}], involving the analogy between water flow and heat flow, a mapping between higher-order structures like \emph{CAUSE\{GREATER-THAN[PRESSURE(beaker), PRESSURE(vial)], FLOW(beaker, vial, water, pipe)\}} will be favored over isolated relations as \emph{GREATER-THAN[DIAMETER(beaker), DIAMETER(vial)]}. 
Since SME is the analogical engine of CogSketch, based on what we established above, we can consider the constraint of systematicity to be automatically extended to the system described in [\cite{cogsketch}], this time within the Visual-spatial domain. When users sketch configurations of objects or geometric transformations, the system translates them into structured relational descriptions, which are then mapped via SME.

\subsubsection*{Inferential projection}

In the Structure-Mapping Engine, inferential projection takes place after the construction of a global mapping (gmap) between the base and target domains. Once this structurally consistent mapping is established, SME searches for additional expressions in the base domain that were not part of the initial mapping but whose arguments are aligned with entities in the target. When such expressions are found, SME projects them into the target domain, thereby generating candidate inferences.

An example of this process is provided in the fluid-flow analogy discussed by [\cite{falkenhainerSME}, pp. 20–23]. In the example, the base domain includes a causal relation stating that if the pressure in a beaker is greater than the pressure in a vial, then water flows from the beaker to the vial through a pipe.
Although this relation is not originally available in the target domain, SME establishes a structural mapping between relevant entities and predicates: ‘Beaker’ is aligned with ‘Coffee’, ‘Vial’ with ‘IceCube’, ‘Water’ with ‘Heat’, ‘Pipe’ with ‘Bar’, and ‘PRESSURE’ with ‘TEMPERATURE’. Since all arguments of the original expression are mapped to corresponding entities in the target, SME projects the following relational structure: $$\emph{CAUSE(GREATER(TEMP(Coffee),TEMP(IceCube)), FLOW(Coffee, IceCube, Heat, Bar))}$$ This new expression is added to the target domain as a candidate inference, even though it was not part of the original input.

CogSketch leverages the same mechanism to project unseen spatial relations, causal connections, or transformations into target sketches, even when these were not explicitly drawn or labeled by the user.

\subsubsection*{Categorization and Property Selection}

The principle of Categorization requires that a metaphor be formed by generating a category derived from the vehicle concept, into which the target concept is subsumed. However, neither SME nor CogSketch have been designed to implement this mechanism. SME treats metaphor construction as a special case of structural mapping between predicate representations, and the same applies to CogSketch. As a result, both systems fail to satisfy the Categorization Constraint. 

Property Selection demands that only the salient features of the category derived from the vehicle concept be projected onto the target concept. Although it may appear that the principle is partially satisfied, given its analogical proximity to the Systematicity Constraint in Structure-Mapping Theory (as both aim to avoid noisy correspondences), strictly speaking, SME and CogSketch do not implement any mechanism for Property Selection as defined within the Metaphor-as-Categorization Theory. Thus, once again, SME and CogSketch exhibit the same limitations with respect to this principle.

\subsection{\textbf{MET}\textsuperscript{CL}}

\subsubsection*{One-to-one mapping, Parallel Connectivity, Systematicity}

\textbf{MET}\textsuperscript{CL} [\cite{lieto2025delta}] is not grounded in SMT. Rather, it adopts a Categorization Theory of Metaphor perspective [\cite{glucksberg2001understanding}], combined with Typicality-based Compositional Logic (TCL) [\cite{lieto2020description}]. According to this framework, metaphors are understood as the combination of elements from the source and target domains into a new ad hoc conceptual category rather than as structural correspondences between two domains. For this reason, \textbf{MET}\textsuperscript{CL} does not represent metaphorical interpretation as a mapping procedure, nor does it define source-target domains in SMT's sense. Instead, it creates the metaphor by computing a blended conceptual space in which the combined prototype inherits the salient properties of the source concept, and reconciles them with the existing prototype of the target. For reasons intrinsic to the system's design, \textbf{MET}\textsuperscript{CL} does not implement any of the constraints derived from SMT that require predicative structures and domain-domain or argument-argument mappings. The system does not comply with one-to-one mapping, parallel connectivity, nor systematicity.

\subsubsection*{Inferential projection}

In the context of Structure-Mapping Theory (SMT), inferential projection refers to the ability of an analogical mapping to produce new inferences in the target domain. This corresponds to the generation of new predicates, attributes, or properties in the target that are not explicitly present in the original data. \textbf{MET}\textsuperscript{CL} generates novel representations that include properties not explicitly present in the target, emerging from the combination of concepts. These representations have been judged as informative by human evaluators, who assigned positive ratings to the generated properties [\cite{lieto2025delta}]. For these reasons, \textbf{MET}\textsuperscript{CL} satisfies the Inferential Projection constraint.

\subsubsection*{Categorization and Property Selection}

\textbf{MET}\textsuperscript{CL} explicitly assumes the Categorization constraint, as it is a foundational assumption of the "metaphor as categorization" paradigm on which the system is based. The theoretical core of the system is the idea that understanding a metaphor involves categorizing the target as a figurative member of the class represented by the source. \textbf{MET}\textsuperscript{CL} constructs a shared category between two concepts $C_1$ and $C_2$ through the HEAD-MODIFIER mechanism: one of the two is selected as the HEAD (providing the main category—or the ontological reference for the resulting concept), while the other acts as the MODIFIER (altering the meaning of the HEAD by introducing additional properties, according to a typicality gradient). Moreover, as prescribed by the Property Selection constraint, the target $C_H$ (HEAD) is enriched with prototypical properties from the source $C_M$ (MODIFIER) only if these are compatible. For the purpose of implementing this constraint, property inheritance must be filtered and contextually grounded: in this respect as well, \textbf{MET}\textsuperscript{CL} satisfies the principle, as it is grounded in TCL logic [\cite{lieto2018description}], which was specifically designed to handle typical properties annotated with degrees of belief. As a result, only a compatible subset of the source’s properties is actually included in the combined representation. 

\subsection{LLMs}

\subsubsection*{One-to-one mapping and Parallel connectivity}

Large Language Models learn to represent and generate language by mapping tokens to high-dimensional vector spaces (embeddings). Embeddings and model parameters are optimized through self-supervised training objectives, such as next-token prediction in autoregressive models. According to Gentner’s Structure Mapping Theory, analogical reasoning requires systematic, bijective correlations between elements of structured representations across the source and target domains. LLMs, on the other hand, rely on high-dimensional, distributed representations where relational mappings are not constrained, and possibly many-to-many. 

This gap has observable consequences. For example, evaluation under distractor-rich conditions reveals critical limitations, even though LLMs frequently achieve high accuracy on analogy tasks, including standard four-term analogies (A:B::C:?) [\cite{analogygenLLM}, \cite{webb2023emergent}]. The Scientific Analogical Reasoning (SCAR) benchmark [\cite{yuan2023beneath}], which comprises over 400 analogies with semantically confusable distractors, shows that GPT-4 and related models frequently choose structurally incorrect completions, indicating a failure to enforce exclusivity in mapping. Furthermore, \cite{musker2025llmsmodelsanalogicalreasoning} demonstrate that, despite the fact that LLMs' performance may appear to be on approximately the same level as that of humans, response variability increases rapidly when prompts are changed or distractions are introduced: the invariance assumption of bijective mapping, which is essential to SMT, and states that bijective structural correspondences should hold up under surface-level reformulations, is broken by prompt-induced response variability.

These findings also serve as indirect indicators of LLMs’ inability to satisfy the parallel connectivity constraint. Response variability observed in LLMs when prompts are rephrased or distractors are introduced suggests a breakdown in this requirement. If a model alters its analogical mapping based on superficial linguistic reformulations, this indicates that argument-level correspondences are not governed by stable relational roles. Support for this interpretation comes from \cite{stevenson2023large}, who find that while LLMs perform roughly at the same accuracy level as children aged 7-12 on proportional analogies (A:B::C:?), their success is largely driven by associative relations (semantic similarity) between C and candidate D, rather than by mapping the underlying relational structure.  

\subsubsection*{Systematicity}

The systematicity constraint, central to Structure Mapping Theory (SMT), posits that analogical reasoning involves a preference for coherent systems of interrelated predicates over isolated relations. This bias toward nested causal structures in analogy is well documented in human cognition: classic empirical studies demonstrate that individuals, including children, prefer analogies that preserve interconnected systems of relations over those involving isolated attributes [\cite{gentner1986systematicity}, \cite{markman1993structural}, \cite{gentner2006systematicity}, \cite{clement1991systematicity}]. To date, there is no strong experimental evidence demonstrating that Large Language Models satisfy this constraint. The current state of the literature does not allow for a definitive determination of whether LLMs implement the systematicity constraint as formulated within SMT. The matter remains open to debate. For instance, in their examination of story analogies, \cite{webb2023emergent} observe that GPT-3 seems to capture the systematicity effect: the model appears to be sensitive to deeper relational structures in analogical reasoning, just like human participants. The research has been criticized by \cite{lewis2024evaluating} and \cite{hodel2023response}, who raise methodological and interpretive concerns.

Similarly, \cite{musker2024semantic} find that LLMs such as GPT-4 and Claude-3 Opus achieve human-level performances on a variety of analogical tasks. However, based on analyses of model sensitivity to semantic perturbations and distractors, the authors conclude that the fundamental mechanisms underlying analogical reasoning in LLMs, even in the most sophisticated models, differ from those employed by humans. 

Because of the current lack of empirical certainty and notwithstanding the growing evidence that LLMs do not follow the systematicity constraint as described in SMT, it would be prudent to avoid establishing definitive conclusions. Even though evidence tends to support the idea that LLMs fail to display a built-in bias toward relational coherence and hierarchical structure, some isolated findings of apparent systematicity suggest that we should be cautious. We can provide a preliminary partial functional score of 1 for LLM's systematicity.

\subsubsection*{Inferential Projection}

Large Language Models do not implement inferential projection as defined in Structure-Mapping Theory, insofar as it refers to the capacity of a system to generate novel expressions in the target domain by projecting relational structures from the base domain. LLMs do not build explicit mappings across domains, nor do they encode symbolic predicate-argument structures. However, as argued in the previous section, the implementation of this constraint must be assessed in broader terms, taking into account the possible emergence of behaviors functionally aligned with the principle. From this standpoint, it is evident that LLMs exhibit behaviors that are equivalent to inferential projection: they generalize relational patterns from base to target contexts and produce novel appropriate results based on those patterns.
A growing body of research, including \cite{webb2023emergent}, \cite{ye2024analobench}, \cite{ding2023fluid}, and \cite{musker2024semantic}, although with variable results, demonstrates that LLMs excel in abstract and symbolic analogies (e.g., matrix completion, sequence induction) and, although with progressively decreasing performance, are capable of performing analogical reasoning in narrative, textual, and even visual tasks. 

\subsubsection*{Categorization and Property Selection}

The core assumption of the Categorization Theory of Metaphor (CTM) is that metaphors operate by interpreting the vehicle concept as the label of an ad hoc category, an on-the-fly conceptual group whose members share a set of contextually relevant features. This process also involves selecting only salient properties of the vehicle to include in the shared category. Whether Large Language Models, which lack explicit representations of categories, engage in metaphor generation in a way that adheres to the theoretical constraints of CTM remains an open and unresolved question. Recent empirical work has begun to explore this possibility, albeit without offering definitive answers. \cite{kim2023metaphorian} developed \textit{Metaphorian}, a system based on GPT-3 tasked with generating scientific metaphors. Their study shows that, under appropriate prompting conditions, the model can produce metaphors that exhibit cross-domain property mappings and context-sensitive relevance. These outputs may suggest some degree of functional alignment with the CTM framework, but the mechanisms underlying such outputs remain opaque: the study does not test whether the model is performing any form of feature selection or category construction in the CTM sense, nor does it isolate the cognitive-like components involved.

\cite{wachowiak-gromann-2023-gpt} assessed GPT-3's ability to determine the proper source domain of metaphorical expressions in the absence of pre-established category inventories. These findings might suggest that the model could extract significant attributes from the metaphor vehicle and relate them to the topic, since accurate domain inference requires access to salient semantic features and conceptual clusters.

When combined, these studies offer some early evidence that LLMs might be able to recognize important cross-domain similarities and apply them to tasks involving metaphors, but current evidence is not sufficient to show that these abilities represent the type of context-sensitive categorization processes that CTM suggests. Rather than requiring the creation of new conceptual groupings, the observed behaviors could simply be the consequence of statistical generalizations over large training corpora. Distinguishing between these possibilities remains a key challenge for future research.

\subsection{Results for $\boldsymbol{FSR}_\mathcal{M}$}
\begin{table}[H]
\centering
\resizebox{\textwidth}{!}{%
\begin{tabular}{l|cccccccccccc|cc|c}
\toprule
\multicolumn{1}{c|}{\raisebox{-1.7ex}{$\mathcal{M}$}}
      & \multicolumn{2}{c}{\makecell{$C_1$}} 
      & \multicolumn{2}{c}{\makecell{$C_2$}} 
      & \multicolumn{2}{c}{$C_3$} 
      & \multicolumn{2}{c}{\makecell{$C_4$}} 
      & \multicolumn{2}{c}{\makecell{$C_5$}} 
      & \multicolumn{2}{c}{\makecell{$C_6$}} 
      & \raisebox{-1.7ex}{$F_\mathcal{M}$} 
& \raisebox{-1.7ex}{$S_\mathcal{M}$} 
& \raisebox{-1.7ex}{\textbf{FSR}$_\mathcal{M}$} \\

 {\scriptsize}
      & \multicolumn{2}{c}{{\scriptsize $w=0.1$}}
      & \multicolumn{2}{c}{{\scriptsize $w=0.1$}}
      & \multicolumn{2}{c}{{\scriptsize $w=0.3$}}
      & \multicolumn{2}{c}{{\scriptsize $w=0.1$}}
      & \multicolumn{2}{c}{{\scriptsize $w=0.3$}}
      & \multicolumn{2}{c}{{\scriptsize $w=0.1$}}
      & & & \\
\specialrule{0.4pt}{0pt}{0pt}
 & $f_i$ & $s_i$ & $f_i$ & $s_i$ & $f_i$ & $s_i$ & $f_i$ & $s_i$ & $f_i$ & $s_i$ & $f_i$ & $s_i$ & & & \\
\textbf{CogSketch} & 0 & 1 & 0 & 1 & 0 & 1 & 0 & 1 & 1 & 0 & 1 & 0 & 0.4 & 0.6 & \textbf{0.65} \\
\textbf{SME} & 0 & 1 & 0 & 1 & 0 & 1 & 0 & 1 & 1 & 0 & 1 & 0 & 0.4 & 0.6 & \textbf{0.65}  \\
\textbf{MET}\textsuperscript{CL} & 1 & 0 & 1 & 0 & 1 & 0 & 1 & 0 & 0 & 1 & 0 & 1 & 0.6 & 0.4 & \textbf{1.46} \\
\textbf{LLMs} & 1 & 0 & 1 & 0 & 1 & 0 & 0 & 1 & 1 & 0 & 1 & 0 & 0.9 & 0.1 & \textbf{8.18} \\
\bottomrule
\end{tabular}
}
\captionsetup{width=0.9\textwidth}
\caption{Functional ($\boldsymbol{F_\mathcal{M}}$) and structural ($\boldsymbol{S_\mathcal{M}}$) scores with final Functional/Structural Ratio ($\boldsymbol{FSR_\mathcal{M}}$). Constraints: {$C_1$} = 1:1 Mapping, {$C_2$} = Parallel Connectivity, {$C_3$} = Systematicity, {$C_4$} = Inferential Projection, {$C_5$} = Categorization, {$C_6$} = Property Selection. The weights reflect the theoretical relevance of each constraint: $C_3$ and $C_5$ are assigned higher weights ($w = 0.3$), while all others are equally weighted ($w = 0.1$).}
\label{tab:FSR}
\end{table}

Since the raw $FSR_\mathcal{M}$ values are not bounded in $[0,1]$, and lower $FSR_\mathcal{M}$ values indicate stronger structural plausibility, in order to maintain consistency with the other components, we compute $FSR'_{\mathcal{M}}$ by inverting $FSR_M$:

\begin{equation}
I(FSR_{\mathcal{M}})= \frac{1}{FSR_{\mathcal{M}}}
\end{equation}

\noindent So that higher values are guaranteed to correlate with a higher structural score. We can then use a normalization formula to map these values into the $[0,1]$ interval:

\begin{equation}
FSR'_{\mathcal{M}} = \frac{I(FSR_{\mathcal{M}})}{1+ I(FSR_{\mathcal{M}})}
\end{equation}

\noindent
The result in $[0,1]$ can be interpreted as a structurality index: the more closely $FSR'_{\mathcal{M}}$ approaches $1$, the more strongly the model structurally aligns with the cognitive theoretical framework of reference.
In Tab. 2, we compare the proposed metric with the linear alternative, where the final score is directly given by $S_M$. Since \(F_M=1-S_M\), any linear combination $$\alpha F_M+\beta S_M $$ reduces to $$\alpha + (\beta-\alpha)S_M.$$
The final score is thus determined by $S_M$.
The results show that, after normalization, differences between the two formulations remain negligible.

\begin{table}[H]
\centering
\begin{tabular}{c|cccc}
\toprule
 & \textbf{CogSketch} & \textbf{SME} & \textbf{MET}\textsuperscript{CL} & \textbf{LLMs} \\
\midrule
\textbf{Non-linear} & 0.606 & 0.606 & 0.407 & 0.109 \\
\textbf{Linear}     & 0.600 & 0.600 & 0.400 & 0.100 \\
\bottomrule
\end{tabular}
\captionsetup{width=0.9\textwidth}
\caption{Comparison between non-linear and linear normalized Functional/Structural scores (\(FSR'_{\mathcal M}\)).}
\label{tab:FSR_comparison}
\end{table}

\subsection{Sensitivity analysis for $\boldsymbol{FSR}_\mathcal{M}$}

To assess the robustness of the proposed $FSR_\mathcal{M}$ metric with respect to variations in the weighting $\{w_1, \dots, w_k\}$ of the structural constraints $\{c_1, \dots, c_k\}$, a local sensitivity analysis was performed.
The analysis was designed to estimate how sensitive each model’s $FSR_\mathcal{M}$ value is to changes in $\{w_1, \dots, w_k\}$, and whether the overall ranking of models remains consistent when the weights vary within plausible ranges.
We employed a local one-at-a-time (OAT) sensitivity analysis: in this procedure, one weight $w_i$ at a time is perturbed by a fixed percentage (\(+30\%\), \(-30\%\)), while the remaining weights are renormalized proportionally:

\begin{equation}
w_j' = \frac{1 - w_i'}{1 - w_i} \, w_j \quad \text{for each } j \neq i
\label{eq:renorm}
\end{equation}

For each perturbed configuration, the $FSR_\mathcal{M}$ values were recalculated for all models.
The variation in $FSR_\mathcal{M}$ relative to the baseline results was then expressed in percentage terms. The resulting $\Delta_i$ values quantify the percentage change in $FSR_\mathcal{M}$ scores when a single weight is increased or decreased.

Across all perturbations of \(\pm30\%\), the $FSR_\mathcal{M}$ ranking remained unaffected, indicating that the metric is not dependent on small variations in the assumed baseline weights (\textbf{Tab. 1}) 

As shown in \textbf{Figure 1}, the largest deviations occur for C3 (Systematicity) and C5 (Categorization). C5 produces the highest sensitivity for SME, CogSketch, and METCL (±36–39\%). C3 also exerts a medium-to-high influence on all models. MET\textsuperscript{CL} is moderately sensitive to C3–C5, and LLMs show the most pronounced response to C4 (Inferential Projection) (±42\%), as this is the only constraint satisfied by LLMs.

\begin{figure}[H]
    \centering
    \includegraphics[width=0.9\textwidth]{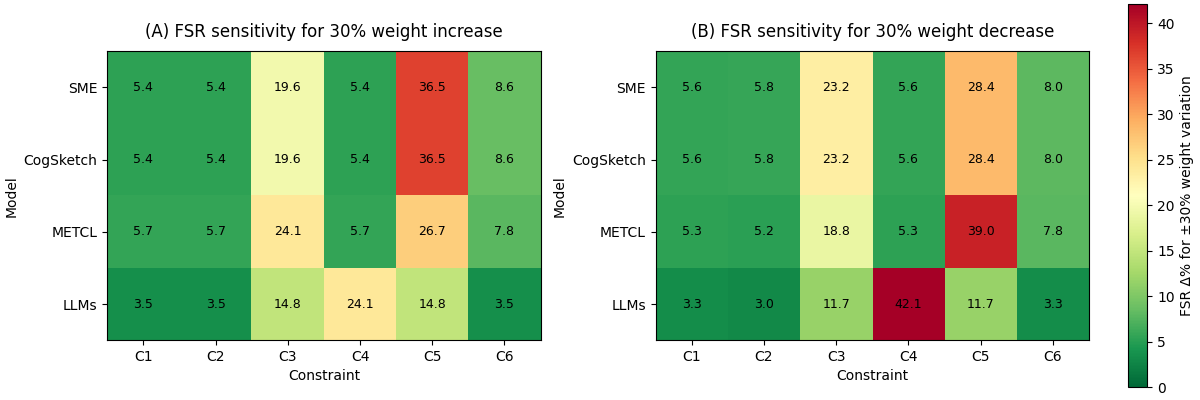}
    \captionsetup{width=0.8\textwidth}
    \caption{Sensitivity of the FSR metric to $\pm 30\%$ variations in constraint weights. 
    Panel (A): positive perturbations; Panel (B): negative perturbations. 
    Color intensity represents the magnitude of percentage change relative to the baseline configuration.}
    \label{fig:sensitivity}
\end{figure}

\section{Generality}

\noindent{Generality is a critical dimension in the evaluation of computational models of cognitive processes. It examines their capacity to replicate human ability in cross-domain tasks, and refers to the breadth of cognitive functions a model can simulate. The higher a model scores in generality, the greater variety of tasks it can execute -  whether in a functional or structural fashion. According to the Minimal Cognitive Grid [\cite{lieto2021cognitive}], models that demonstrate higher generality are considered to offer greater explanatory power, as they simulate multiple facets of human cognition rather than narrow, task-specific operations. 
In the following sections, we will assess the generality of each of the reviewed AI systems (Structure-Mapping Engine, CogSketch, \textbf{MET}\textsuperscript{CL}, and Large Language Models}). 

In order to ground our evaluation of generality in an empirically validated cognitive framework, we selected task domains by drawing from the Cattell–Horn–Carroll (CHC) theory of cognitive abilities [\cite{carroll1993human}, \cite{mcgrew2005cattell}]. 

\noindent The CHC psychometric model is the most thoroughly researched, empirically validated, and comprehensive hierarchical taxonomy of human cognitive abilities, organized into three strata: 1. domain-independent capacities; 2. a middle layer of broad abilities; 3. numerous narrow abilities. Among the most relevant areas, CHC includes: Fluid Intelligence (Gf), Quantitative Knowledge (Gq), Reading-Writing (Grw), Memory (Gsm/Glr), Visual Processing, Auditory, Olfactory, Tactile Processing, Psychomotor and Kinesthetic abilities, etc. (see \textbf{Fig. 1}). 

\begin{figure}
    \centering
    \includegraphics[width=\textwidth]{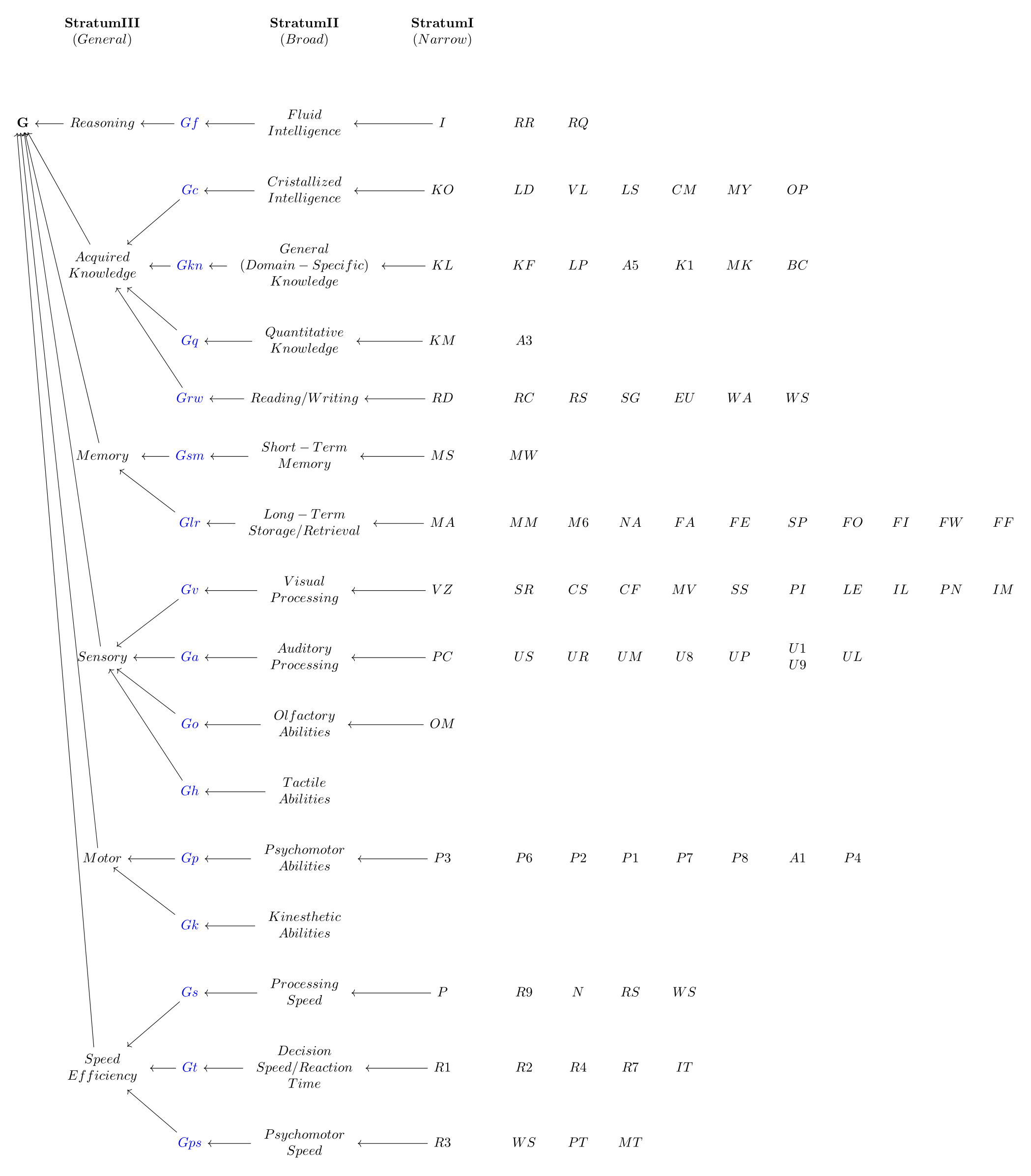}
    \captionsetup{width=0.8\textwidth}
    \caption{A schematic representation of the Cattell-Horn-Carroll taxonomy of human cognitive abilities. The general intelligence factor (\textbf{G}) branches into an array of cognitive/sensory-motor general domains, each further subdivided in a set of broad (\textbf{Stratum II}) and  narrow (\textbf{Stratum I}) abilities.}  
    \label{fig:placeholder}
\end{figure}

Rather than adopting the full taxonomy proposed by the CHC theory, we only retain a subset of five broad task domains with direct grounding in CHC, specifically: 1. \textbf{Quantitative knowledge} [G\emph{q}]; 2. \textbf{Fluid reasoning} [G\emph{f}] (including inductive, analogical and metaphorical reasoning); 3. \textbf{Visual processing} [G\emph{v}]; 4. \textbf{Language and Verbal Knowledge} (an aggregation of G\emph{c}/G\emph{a}/G\emph{rw} covering comprehension, lexical/grammatical competence, and reading/writing tasks); 5. \textbf{Sensory/Motor Abilities} (an ad hoc supercategory formed by collapsing \(G{v}\), \(G{o}\), \(G{h}\), \(G{p}\), and \(G{k}\)).

In line with cognitive and neuropsychological evidence emphasizing the centrality of embodiment in human cognition,
we partition the domains into two macro-categories: 1. Cognitive Abilities (Quantitative, Fluid, Visual, and Language/Verbal Reasoning); 2. Sensory/Motor Abilities. Each macro-category is assigned a weight of 0.5 in the computation of the final generality score ($G_\mathcal{M}$).
The high weight assigned to the Sensory/Motor domain reflects the importance that the literature on embodied cognition acknowledges to it. A substantial body of work in cognitive science argues that higher-level cognition is deeply rooted in perception and action, and that cognitive processes cannot be understood independently of their ability to interact with the physical environment [\cite{barsalou2008grounded, wilson2002six, gallese1996action, clark1999embodiment}].
Under these assumptions and evidences, systems that lack capabilities in the Sensory/Motor domain do not achieve the full range of general cognitive abilities observed in humans.
However, to check how robust our results are, we also provide an alternative scoring scheme where Sensory/Motor is treated as an additional domain with the same weight as the others (Table ~\ref{tab:task_domains}).

Working/short-term memory (\(Gwm\)) and long-term storage/retrieval (\(Glr/Gr\)) are not included, along with Processing speed (\(Gs\)) and decision/reaction time (\(Gt/Gps\)), since they will be captured in the \textit{Performance Match} evaluation.

Each domain has been selected for its central role in human cognitive processing. Together, they form a representative spectrum of cognitive functions that provides a diverse set of testing areas for the systems. While supported by extensive research in cognitive science [\cite{chomsky2011language}, \cite{tversky1977features}, \cite{kintsch1994text}, \cite{gentnermarkman}, \cite{holyoak2012analogy}], the list might be deemed incomplete; it is, in fact, extensible according to empirical evidence. A paramount feature of the CHC theory is that it is not static; rather, it is a dynamic model that is continuously reorganized based on research advancements. 

The capability of a model $\mathcal{M}$ in each of the $n$ task domains $\{ d_1, \dots, d_n\}$ is evaluated on a three-point ordinal scale: for each domain, let $g_j\in\{0, 0.5, 1\}$ denote the score of $\mathcal{M}$ in domain $d_j$. We assign values:

\begin{equation}
g_j =
\begin{cases}
1 & \text{if $\mathcal{M}$ covers the domain } d_j \\
0.5 & \text{if $\mathcal{M}$ partially covers the domain } d_j \\
0 & \text{otherwise}
\end{cases}
\end{equation}

\noindent The overall generality score is computed as:

\begin{equation}
G_\mathcal{M} = 0.5 \cdot \overline{g_{\text{cognitive}}} + 0.5 \cdot g_{\text{sensorimotor}}
\end{equation}

\noindent Where:

\begin{equation}
{\overline{g_{\text{cognitive}}}} = \frac{1}{n}\sum_{i=1}^ng_j 
\end{equation}
\noindent And $g_{\text{sensorimotor}}$ is the score assigned to the Sensory/Motor domain. This yields a value in the range [0,1], where higher scores indicate greater generality, i.e., the ability to operate across a wider array of cognitive functions.

\subsection{Structure-Mapping Engine/CogSketch}\label{s:section}

\subsubsection*{Quantitative Knowledge}

SME and CogSketch lack any architecture for representing or manipulating numerical symbols, discrete or continuous quantities, and arithmetic operations. Neither system includes mechanisms for reasoning about mathematical concepts or operations. Their representational format is entirely based on predicate logic structures that capture qualitative relational information. Even in domains that involve geometrical representations, such as those processed by CogSketch, all comparisons are carried out over discrete symbolic descriptors. For example, relative length, angle, or position are assessed via qualitative heuristics or rule-based categories (e.g., “approximately equal”, “more acute”, “left-of”) rather than through continuous, quantitative modeling. The systems are not capable of executing any task that requires mathematical reasoning, numerical abstraction, arithmetic, geometry, algebra, calculus, etc. 

\subsubsection*{Fluid Reasoning}

SME and CogSketch implement one of the core subcomponents of Fluid Reasoning: analogical mapping and relational inference. As clearly shown in \cite{falkenhainerSME}, analogical relational mapping is embedded in SME’s design structure. The system has been built to explicitly provide matching algorithms consistent with Gentner’s Structure Mapping Theory [\cite{gentnerstructuremapping}]. SME operationalizes the principles of Structure Mapping by emphasizing the alignment of relational structures rather than surface-level features and identifying correspondences between elements from the "base" (B) and "target" (T) domains, based on their shared relational structures. Briefly delving into the Structure-Mapping architecture will help us provide evidence for our claim about its relational mapping capabilities. First of all, SME requires input descriptions of the (B, T) domains expressed in predicate logic. Following SME's authors example, in order to describe a hierarchical representation of the items in (B, T), objects and constants are considered as order-0 items, while predicates' order is one plus the maximum order of its arguments (\textit{e.g.} GREATER-THAN(x, y) is a first order predicate if (x, y) are objects). An item is second-order if it takes a first-order predicate as its argument (\textit{e.g.} CAUSE[GREATER-THAN(x, y), BREAK(x)]. These predicate calculus representations capture \textit{entities}, \textit{predicates} of three different types (\textit{functions, attributes, relations}) and \textit{Dgroups} (a group of descriptions about some facts, including entities, their attributes, and the relations among them). 

SME's algorithm establishes a set of constraints as analogical mapping rules from B to T (\textit{e.g.} let ${B_i}$ and ${T_i}$ indicate items in the base and target representations; let the subsets ${b_i}$ and ${t_i}$ denote objects in B and T; some of the rules are: ${M:b_i}\to {t_i}$ (objects in B are mapped to objects in T); $M: COLLIDE({b_i},  {b_j}) \to  COLLIDE({t_i}, t_j)$ (relations in B are mapped to relations in T), with the overarching rule for mapping being \textit{systematicity}. Given the representation of B and T as Dgroups, SME builds all the consistent match hypotheses, each called \textit{global mapping} or \textit{gmap}, then proceeds computing the candidate inferences and a match evaluation score for each gmap. 

In section 4.2 of their work, the authors give a simplified visualization of the relational mappings performed by SME on a very classic analogy in science (solar system/Rutherford atom). With solar system formally represented in the base dgroups and Rutherford model in the target domain, SME constructed three different interpretations, ranked by Structural Evaluation Score (SES). The best graded one (SES = 6.03) matched the nucleus with the sun and the planet with the electron, while the lowest scoring match hypothesis (SES = 1.87) implied an equivalence between the mass of the sun and that of the electron.
In addition to providing evidence of the structural mapping capabilities of the system, the example demonstrates a crucial aspect of the interpretation of analogy in terms of structure-mapping. 

\noindent
Considered that SME serves as the engine of CogSketch's [\cite{forbus2008cogsketch}] analogical reasoning capabilities, hence what we have here stated about the relational mapping powers of the Structure Mapping Engine, also applies to the latter system.

Despite its strength in analogical reasoning, SME does not support several other abilities under the task domain as defined, for example, in \cite{otero2017brief}.

\subsubsection*{Visual processing}

As seen, SME's only explicit (by design) capabilities lie in: a) mapping relational similarities between two structured representations of a base and a target domain, and b) assigning a structural evaluation score (SES) for each candidate inference. When assessing SME's visual-spatial reasoning power as a standalone tool, we must therefore keep in mind that its specialization does not allow for anything else than building structural mapping hypotheses and measuring their accuracy. However, its very limitation is the cause for its flexibility when broadened into a more general system such as CogSketch: while SME provides the analogical reasoning framework, CogSketch bridges visual information and structured symbolic reasoning so that, together, they can perform a variety of visual-spatial reasoning tasks, including understanding sketches and spatial relations, comparing shapes and solving Raven's Progressive Matrices. 

A relevant working hypothesis proposed in the works of Forbus, Gentner, Lovett, and others [\cite{lovett2017modeling}] suggests that structure-mapping processes are central to visual reasoning. Specifically, perceptual processing transforms continuous properties into qualitative spatial representations by quantizing them into discrete, meaningful units. These representations can then be manipulated symbolically to facilitate reasoning. Supported by psychological evidence from \cite{lovett2009using}, this hypothesis establishes a fundamental link between analogy-making and perceptual processing. It has also enabled the generalization of computational models such as SME into the domain of visual-spatial reasoning, in a cognitively plausible manner.

As shown by \cite{lovett2007analogy}, when employed with the spatial information provided by sKEA [\cite{forbus2005structure}] (CogSketch's precursor), SME is capable of solving a set of  problems in the form of 2x2 and 3x3 Raven's Progressive Matrices. In order to accomplish the goal, it sufficed to encode the matrices in the form of geometric analogy problems (A:B::C:?), making SME  detect similarities and differences between representations populated by, for instance, "top-left entry", "bottom-left entry", "top-right entry" etc., where each entry is a qualitative representation of shapes and positions produced by sKEA. Making SME run comparisons between individual entries in the matrix (e.g. top-right/bottom-right) and build a representation of the differences between the two entries $ \Delta({e_1}, {e_2})$ is the first stage of a twofold process. The second one builds a comparison between the two previously generated {$\Delta$}, so that \textit{$\Delta$(upper-left, lower-left)} is compared to \textit{$\Delta$(upper-right, answer)} for each of the six possible answers. The correct answer, eventually, will be the one whose $\Delta$ is most similar to \textit{$\Delta$(upper-left, lower-left)}, representing the known answer. 
The process was tested on two out of the five sections of the Standard Progressive Matrices benchmark, making SME correctly answer 22 out of 24 problems and allowing the authors to compare the system's performance to human performance in the same task domain. SME has also been used in AI models designed to solve physical reasoning problems [\cite{klenk2005solving}]. 

Solving tasks like Raven’s Progressive Matrices involves identifying critical differences in geometric features such as scale changes, orientation, reflection, and rotation. This process can be considered as a proxy for assessing spatial reasoning capabilities.

\subsubsection*{Language and Verbal Knowledge}

Language processing represents a critical domain for generality in computational systems: it challenges models to understand, reason about, and manipulate the semantic and syntactic structures embedded in natural language. Within this domain, benchmarks such as GLUE (General Language Understanding Evaluation, \cite{Glue}) and SuperGLUE [\cite{superGlue}] serve as standard frameworks for testing capabilities in textual assignments. These benchmarks comprise a broad array of tasks, including natural language inference (NLI), question answering (QA), co-reference resolution, semantic similarity assessment, and more.

The emphasis on text comprehension highlights a critical limitation of systems like SME and CogSketch, which are not designed for NLP tasks.
Tasks like semantic similarity or categorization, which depend on identifying overlapping relational structures between sentences, could align with SME’s capabilities if the input representations were sufficiently robust. Nonetheless, significant limitations arise when SME is applied to text-based tasks. Natural language contains ambiguities, context-dependent semantics, and subtle lexical variations that are challenging to encode in purely relational terms. Tasks such as co-reference resolution, which involve tracking entities and their referents across sentences, or question answering, which demands nuanced integration of context, are inherently beyond SME’s design. As a result, SME is capable of achieving success only in narrowly scoped tasks where input representations explicitly encode relational reasoning. For most text comprehension tasks in GLUE or SuperGLUE, SME’s performance remains negligible.

Similarly, CogSketch lacks any native mechanisms for processing textual input and is thus inherently misaligned with the requirements of text comprehension tasks. Extending CogSketch to such domains would require the development of translation mechanisms that map linguistic descriptions into spatial or relational diagrams. While this approach might hold potential in hybrid tasks where text interacts with spatial reasoning, no research to date has demonstrated such functionality. Moreover, even in these hypothetical scenarios, the challenges of parsing, disambiguation, and context-dependent reasoning persist, making the adaptation speculative at best.
Consequently, CogSketch scores 0 on all tasks within text comprehension benchmarks such as GLUE or SuperGLUE.

\subsection{\textbf{MET}\textsuperscript{CL}}

\subsubsection*{Quantitative Knowledge}

In CHC theory, Quantitative Knowledge encompasses a system’s ability to understand and manipulate numerical expressions, solve arithmetic problems, and interpret quantitative relations. \textbf{MET}\textsuperscript{CL} handles only symbolic–semantic combinations, operating at the level of typical features attribution and prototype combination. As such, it has no conceptual or operational mechanisms for numeric quantity representation, arithmetic operations or mathematical reasoning at large.

\subsubsection*{Fluid Reasoning}

\textbf{MET}\textsuperscript{CL} exhibits noteworthy strengths in several narrow subcomponents of the domain, while remaining limited in other areas of fluid intelligence. Its core mechanism enables the creation of composite conceptual prototypes useful in metaphor generation. This process can be considered to perform inductive generalizations akin to those observed in human categorization processes, as it abstracts typical features from source and target prototypes and blends them in a hybrid concept. These capabilities cover the inductive reasoning area of fluid cognition [\cite{carroll1993human}].

Nonetheless, \textbf{MET}\textsuperscript{CL} misses other critical aspects of fluid intelligence as codified in \cite{mcgrew2005cattell}: it is not a model of abstraction, general reasoning and problem-solving, and cannot be tested on tasks such as Raven's Matrices or the Wechsler Scale, which are typically used to measure fluid intelligence. Since \textbf{MET}\textsuperscript{CL} implements key narrow abilities within the domain (specifically, conceptual blending and, to some degree, inductive reasoning) it earns a partial score of 0.5 for Fluid Reasoning.

\subsubsection*{Visual processing}

\textbf{MET}\textsuperscript{CL} does not engage with visual representations or spatial reasoning in any form.
The system lacks any mechanism for processing or reasoning about visual inputs and offers no interface for perceptual input. The complete absence of visual or spatial reasoning capacities results in a categorical exclusion from the Visual processing domain, for which it receives a score of 0.

\subsubsection*{Language and Verbal Knowledge}

We combine the domains of "Reading and writing" and "Comprehension–Knowledge" within the CHC framework into a more general (and better suited to computational modeling) NLP area. It includes the ability to comprehend, generate and use language, show sensitivity to grammatical rules, understand semantic content, manipulate lexical information, etc. [\cite{carroll1993human, mcgrew2005cattell}].

\textbf{MET}\textsuperscript{CL} exhibits partial coverage of this domain. Despite not being a NLP system in the conventional sense, it nonetheless engages with natural language content through a specialized pipeline developed for metaphor processing: the system interfaces with linguistic data by extracting source and target concepts from metaphorical expressions, as encoded in structured resources such as MetaNet. These concepts are then mapped onto semantic prototypes, enabling the system to perform metaphor classification and generation. In empirical evaluations reported by \cite{Lieto, Pozzato, lieto2025delta}, the system demonstrated the ability to generate metaphorical expressions across datasets such as NN-450 and MetaNet with high coverage (above 98\%) and positive human evaluation ratings (mean score around 6/10 for interpretability and plausibility). Moreover, the system’s outputs improved performance in metaphor classification tasks when used to enhance zero-shot prompting in large language models.
However, its linguistic competence remains narrowly scoped. It does not support open-ended text comprehension, or any other tasks that define the broader NLP environment, for this reason, we assign \textbf{MET}\textsuperscript{CL} a partial score of 0.5 for Language and Verbal Knowledge.

\subsection{Large Language Models}

\subsubsection*{Quantitative Knowledge}

LLMs’ performance in this domain has improved significantly: on the MATH dataset [\cite{hendrycks2021measuring}], GPT-4 achieves over 80\% accuracy with Chain-of-Thought prompting, outperforming prior GPT-3 models that hovered around 40–50\%. These gains are evident across subcategories including arithmetic, algebra, number theory, and geometry, though they remain inconsistent in multi-hop logic and long-derivation problems. On GSM8K [\cite{zhong2024achieving}], GPT-4 with the "Deeply Understanding the Problems" (DUP) prompting technique achieves a new SOTA result, with an accuracy of 97.1\% under the zero-shot setting. 

Although their performance varies depending on input complexity, length of reasoning chains, and prompt quality, the mere execution of tasks across this spectrum supports the claim that LLMs possess domain-relevant functionality. For this reason, and despite acknowledged limitations, we consider mathematical reasoning a domain that is functionally covered by Large Language Models.

\subsubsection*{Fluid Reasoning}

Abstraction is a core component of fluid intelligence in the Cattell–Horn–Carroll framework. In a recent study conducted by \cite{ARC-LLMs}, LLMs were evaluated using the Abstraction and Reasoning Corpus (ARC) [\cite{chollet2025arc}], a benchmark explicitly designed to test the ability to generalize from very limited examples using human-like inductive abstraction. ARC tasks involve transformations over 2D grids that require identifying relations such as symmetry, sequencing, rotation, containment, and color-based logical rules. Unlike typical language-based tasks, ARC demands domain-general reasoning without linguistic scaffolding.

The findings in the paper show that LLMs—specifically GPT-4 and Claude 2.1—achieve extremely limited success in these abstraction tasks. In zero-shot and few-shot settings, GPT-4 and Claude 2.1 performed significantly below human baselines, solving fewer than 4 out of 40 tasks in the ARC-Challenge subset. This performance is contrasted with the median of 15 tasks solved by unaided human participants (N=118) and 25–27 tasks when humans are given an ARC-specific symbolic language to express their reasoning steps.
Accordingly, LLMs do not fulfill the core criteria of abstract reasoning as defined in cognitive science. 
On the other hand, recent research [\cite{webb2023emergent}; \cite{kojima2022large}] shows that Large Language Models can perform analogy tasks in a variety of different forms, although they are structurally limited in their ability to perform explicit relational mapping as defined in Structure-Mapping Theory (Gentner, 1983). As previously explained, structure mapping requires aligning higher-order relations between entities in distinct knowledge domains (B, T), a task that is traditionally suited to symbolic AI frameworks like the Structure-Mapping Engine [\cite{falkenhainerSME}]. In contrast, LLMs operate primarily through statistical pattern recognition, lacking a mechanism to directly encode or reason over relational structures.

Nonetheless, LLMs have demonstrated a clear functional ability to capture relational patterns: \cite{webb2023emergent} provided evidence in support of the emergence of analogical reasoning capabilities in LLMs. They evaluated OpenAI's influential GPT-3 model [\cite{brown2020}] on a span of zero/few-shot analogical reasoning tasks, also comparing the system's performances with human results. GPT was tested on: 

\begin{enumerate}

    \item \textit{Digit Matrices}: a novel text matrix task based on Raven's Standard Progressive Matrices (Raven, 1938), which consisted of \textit{a}) digit transformations, or \textit{b}) logic problems based on the set relations $OR, AND$ and $XOR$.
    On a N=40 set of human participants, GPT-3 (\textit{text-davinci-003} model variant) outperformed the average human on all problem types (although some participants surpassed the model). It must be noted, anyway, that Digit Matrices - differently from  raw traditional (visual) RPMs - do not require parsing pixels into objects or disentangling visual features and offer GPT process-ready text input.
    \item\textit{Letter String Analogies} (\cite{hofstadter1994copycat}). Both GPT and humans participants had to solve problems as: 
    
    $$a, b, c, d \to a, b, c, e$$ 
    $$d, e, f, g \to d, e, f, ?$$

    varied across a diverse range of single/group letter transformations (extended sequence, successor, predecessor, remove redundant letter etc.) or generalizations (letters to numbers, grouping, reverse order, larger intervals etc.). GPT scored better than human participants on average.
    \item\textit{Four-term verbal analogies}: analogy problems in the already mentioned form of $$A:B::C:?$$ with multiple choice for D, where each letter is a real-world concept. Once again, GPT's average score surpassed that of the participants on three of the four datasets (UCLA VAT, Sternberg and Nigro, SAT, Jones \textit{et al.}). 
    \item\textit{Story Analogies}. GPT was tested on a set of complex story analogies from Gentner, Rattermann and Forbus (1993), in order to evaluate its higher-order mapping capabilities, given that such higher-order relations are central to the standard cognitive theory of analogy (Gentner, 1983). This was the only task domain in which human participants outperformed the model, while a likewise preliminary investigation on GPT-4 displays a more robust performance (OpenAI, 2023). 
    \item\textit{Analogical problem-solving}. A preliminary analysis with partial results reported in (Gick and Holyoak, 1980). 
    
\end{enumerate}

As previously shown, \cite{wachowiak-gromann-2023-gpt} also prove that GPT-3 can infer appropriate metaphorical source domains from target concepts with ~65\% accuracy, without relying on predefined taxonomies, suggesting that LLMs can approximate key mechanisms of metaphorical abstraction, including conceptual blending, and identify shared attributes and semantic clusters across domains. 

\subsubsection*{Visual-Spatial Reasoning}

Visual-spatial reasoning in Large Language Models has emerged as a significant frontier in Artificial Intelligence research. Although traditional unimodal text-based LLMs like GPT-3 demonstrate clear limitations in addressing spatial or geometric tasks due to the lack of visual grounding, multimodal Large Language Models (MLLMs), and large vision-language models (VLMs) in particular, show increasingly strong image-recognition and visual-spatial reasoning abilities. Models such as GPT-4V [\cite{openai2024gpt4technicalreport}] and Gemini Pro V [\cite{geminiteam2024geminifamilyhighlycapable}], which integrate both language and visual processing, have demonstrated impressive performance on visual question answering (VQA), image captioning, and other vision-language tasks when tested on novel multimodal-specific benchmarks such as MMBench [\cite{mmbench}], SEEDBench [\cite{li2023seed}], LAMM [\cite{yin2024lamm}], MME [\cite{fu2024mme}], and LVLM-eHub [\cite{xu2024lvlm}].

These benchmarks evaluate how well models can generate textual responses based on multimodal inputs, including tasks that require visual-spatial reasoning abilities, such as diagram interpretation, spatial layout analysis, geometric transformations, object arrangement, and contextual understanding of visual scenes. The results indicate that models like GPT-4V excel at simpler visual tasks (e.g., identifying objects or answering direct questions about spatial relationships), while they can struggle with tasks that require more sophisticated, abstract reasoning [\cite{li2023evaluating}]. While displaying significant performance degradation in complex scenes compared to simpler tasks, multimodal LLMs can still engage in visual-spatial reasoning tasks.

\subsubsection*{Language and Verbal Knowledge}

Language processing is the primary domain of LLMs, and their performance in this area has been extensively validated across a wide span of benchmarks, including the aforementioned GLUE [\cite{Glue}] and SuperGLUE [\cite{superGlue}]. These models achieve state-of-the-art results on a great variety of tasks related to natural language processing (NLP).  

\cite{brown2020} provided evidence of GPT-3's results in language modeling in a zero-, one-, and few-shot fashion on: 1. LAMBADA [\cite{paperno2016}]: last word prediction in a long range semantic dependency environment. High accuracy (86.4\%) in the few-shot setting and substantial increase from the previous state of the art; 2. HellaSwag, [\cite{hellaswag}]: choosing the best ending to a story or set of instructions, with adversarially mined examples in order to make the task more complicated for the model (79.3\% accuracy); 3. StoryCloze [\cite{mostafazadeh2016corpus}]: selecting the correct ending sentence in a long story (87.7\% accuracy); 4. Closed Book Question Answering: answering questions on broad factual knowledge (TriviaQA, Natural Questions, Web Questions = 71.2\%, 29.9\%, 41.5\% accuracy respectively); 5. Translation; 6. Winograd Schemas Challenge [\cite{winograd}]: determining which word a grammatically ambiguous pronoun refers to (88.6\% accuracy); 7. Common sense  reasoning in PhysicalQA, ARC, UnifiedQA, OpenBookQA (mixed results and inconsistent gains); 8. Reading comprehension, 9. SuperGlue; 10. Natural Language Inference (NLI); 11. Word scrambling and manipulation tasks (anagrams, reversed words, \textit{etc.}); 12. SAT Analogies (65.2\% accuracy, with average score among human applicants of 57\%); 13. News article generation: human accuracy in identifying whether text is model-generated was quantified, showing a clear decrease (i.e., correctly detecting machine generated output becomes more difficult) while the model grows in scale (76\% human accuracy for GPT-3 Small versus 52\% for GPT-3 175B, barely above chance).
LLMs have marked a major step forward in the Natural Language Processing research field as a whole: they consistently outperform earlier approaches, demonstrating their ability to extract and integrate information from complex textual inputs and excelling in many areas of text understanding and generation.

\subsubsection*{Sensory/Motor Abilities}

None of the evaluated models exhibit measurable coverage of the \textbf{Sensory/Motor Domain}, which comprises \textit{tactile} (\(G{h}\)), \textit{olfactory} (\(G{o}\)), \textit{kinesthetic} (\(G{k}\)), and \textit{psychomotor} (\(G{p}\)) abilities---that is, the integration of sensory input and motor control. 
Because all analyzed systems are \textit{disembodied computational architectures}, lacking perceptual interfaces and physical effectors, they cannot instantiate the causal coupling between perception and action that defines this domain. 
Accordingly, all models score \(g_{\text{Sensory/Motor}} = 0\).

\subsection*{Results comparison for Generality}

The following table summarizes the results of the evaluation of generality for the systems considered in our analysis. LLMs obtained the highest score, followed by CogSketch, $\textbf{MET}\textsuperscript{CL}$ and SME. 


\begin{table}[H]
\centering
\footnotesize
\begin{tabular*}{\textwidth}{l||@{\extracolsep{\fill}}c c c c c|c|c}
\toprule
\multicolumn{1}{c}{\raisebox{-1.7ex}{\textbf{Model}}} 
& \multicolumn{5}{c}{\textbf{Task Domain}} 
& \raisebox{-1.7ex}{$\boldsymbol{G_\mathcal{M}}$}
& \raisebox{-1.7ex}{$\boldsymbol{G_\mathcal{M}(1)}$} \\

\cmidrule(lr){2-6}
 & {\scriptsize Quant. Know.} 
 & {\scriptsize Fluid Reas.} 
 & {\scriptsize Vis. Proc.} 
 & {\scriptsize Lan.\&Verb.} 
 & {\scriptsize Sens./Mot.} 
 & & \\

\midrule
\textbf{CogSketch}   & 0 & 0.5 & 0.5 & 0  & 0 & \textbf{0.125} & \textbf{0.200} \\
\textbf{SME}         & 0 & 0.5 & 0 & 0   & 0 & \textbf{0.063} & \textbf{0.100} \\
\textbf{MET}\textsuperscript{CL} & 0 & 0.5 & 0 & 0.5 & 0 & \textbf{0.125} & \textbf{0.200} \\
\textbf{LLMs}        & 1 & 1   & 1 & 1   & 0 & \textbf{0.500} & \textbf{0.800} \\

\bottomrule
\end{tabular*}

\captionsetup{width=0.9\textwidth}
\caption{Comparison of the models across four cognitive domains and one sensory/motor domain. 
The generality index \( \boldsymbol{G_\mathcal{M}} \) is obtained by averaging the cognitive domain scores and combining this mean with the sensory/motor domain score using equal (50/50) weighting. The alternative index \( \boldsymbol{G_\mathcal{M}(1)} \) considers sensory/motor as an additional domain with the same weights as the others (1/5).}
\label{tab:task_domains}
\end{table}

\section{Performance Match}

In this section, we evaluate each model according to its performance in analogy and metaphor.
As introduced by \cite{lieto2021cognitive}, the Performance Match assessment should include, in addition to the evaluation of accuracy, an analysis of error patterns and task execution times\footnote{The idea of comparing execution times in humans and machines is not new in the field of computational cognitive modeling. Despite the architectural differences between the biological and artificial systems that execute the same task, the basic assumption is that execution times provide an indirect, functionalist hint — according to the sense of functionalism adopted in this paper — about the computational mechanisms and processes involved in the task at hand. For example, in commonsense reasoning, the fact that most symbolic- and logic-based AI approaches are computationally intractable, whereas humans perform the same tasks in a very short time, indirectly suggests that the underlying mechanisms adopted by biological and artificial systems are different. Importantly, since Allen Newell's proposal on the time scales of human actions [\cite{newell1994unified}], and more recently within both the Psychometric AI movement [\cite{bringsjord2011psychometric}], which advocates the use of a battery of validated tests to assess the effective “match” between artificial systems and human beings, and the Resource-Rationality framework [\cite{lieder2020resource}], this measure has been considered an important aspect of human-machine comparison.}. 
For each system, we will quantify the alignment between models and human subjects on selected benchmarks. The Performance Match for each model will be calculated as the sum of three different sub-metrics: a percentage delta from the baseline given by human performances, error patterns, and response times. Due to a lack of data from dedicated studies, quantitative measurements will not always be achievable, even though qualitative analyses will be attempted whenever possible. More formally, the aggregate Performance Match score ($PM_\mathcal{M}$) will be computed as the weighted sum: 
\begin{equation}
PM_\mathcal{M} = \alpha \cdot A_\mathcal{M} + \beta \cdot E_\mathcal{M} + \gamma \cdot T_\mathcal{M}, \quad   \text{with} \quad \alpha + \beta + \gamma= 1,\quad \alpha, \beta, \gamma \in [0, 1]
\end{equation}

\noindent
For \textit{Accuracy} ($A_\mathcal{M}$), let $\mathcal{B}=\{B_1, \dots, B_n\}$ denote the set of benchmarks on which a given model is evaluated. For each benchmark $B_i \in \mathcal{B}$, we define a $Delta$ as difference (measured in percentage points) between the model's accuracy ($A_{m,i}$) and the human baseline value ($A_{h,i}$)
\begin{equation}
\Delta_i = A_{{m},i} - A_{{h},i}
\end{equation}

\noindent
The mean deviation from the human baseline across all $n$ benchmarks is then computed as:
\begin{equation}
\bar{\Delta} = \frac{1}{n} \sum_{i=1}^{n} \Delta_i
\end{equation}

\noindent
Then, we compute the Accuracy score ($A_\mathcal{M}$) as follows:

\begin{equation}
A_\mathcal{M}=\frac{1}{1+ |\bar{\Delta}|}
\end{equation}

\noindent
This returns a maximum value of 1 if and only if the model replicates human performance on average, and decreases symmetrically as the deviation grows, whether positive or negative.

\noindent
To account for qualitative aspects of model performance, for each benchmark $B_i$ we define a binary Error Pattern indicator $E_i \in \{-1,+1\}$ based on whether the model replicates typical human error distributions. We assign:
\begin{equation}
E_i =
\begin{cases}
+1 & \text{if $\mathcal{M}$ replicates human error patterns} \\
-1 & \text{otherwise}
\end{cases}
\end{equation}

\noindent
This penalizes systems that diverge from human failure patterns and enables a more informative scoring, especially when data on such patterns is unavailable. Then, we aggregate these across $n$ benchmarks as the mean:
\begin{equation}
\bar{E}=\frac{1}{n}\sum_{i=1}^nE_i  \quad
\end{equation}

\noindent
This results in a score $\bar{E} \in [-1,1]$. Since it is not directly comparable to other normalized scores in $[0, 1]$, we apply a transformation to map $\bar{E}$ onto the unit interval:
\begin{equation}
E_\mathcal{M} = \frac{\bar{E} + 1}{2} \quad \quad E_\mathcal{M} \in [0,1]
\end{equation}

\noindent
A score of 1 indicates reproduction of human-like errors across all benchmarks, whereas a score of 0 indicates a complete failure to replicate any human error pattern. Benchmarks for which no error data is available ($E_i = 0$) do not contribute to the score, avoiding artificial inflation or deflation of the final value.
$E_\mathcal{M}$ is intended to capture the extent to which a model reproduces the \textit{statistical structure} of human failures on a given benchmark. Accordingly, a model is not rewarded or penalized for a single incorrect answer taken in isolation. For example, in a task such as Raven’s Progressive Matrices, the relevant comparison concerns whether the model’s failures cluster around the same matrices that are systematically difficult for humans.

The third component of the \textit{Performance Match} metric captures the extent to which the model replicates the response times of human cognition.
Let $t_{m,i}$ be the model's average response time on benchmark $B_i$,  and $t_{h,i}$ be the average human response time on the same benchmark. The relative temporal deviation for each benchmark is defined as:

\begin{equation}
d_i = \left| \frac{t_{m,i} - t_{h,i}}{t_{h,i}} \right|
\end{equation}

\noindent
This value expresses how far the model's response latency deviates from the human baseline in proportional terms, irrespective of whether the model is faster or slower. To convert this deviation into a similarity score bounded in the interval $(0, 1]$, we define:

\begin{equation}
T_i = \frac{1}{1 + d_i} = \frac{1}{1 + \left| \frac{t_{m,i} - t_{h,i}}{t_{h,i}} \right|}
\end{equation}

\noindent
A value of $T_i = 1$ indicates perfect temporal alignment ($t_{m,i} = t_{h,i}$), while lower values represent increasing misalignment.

\noindent
The overall \textit{Execution Time Score} is computed as the arithmetic mean of the individual scores:

\begin{equation}
{T_\mathcal{M}} = \frac{1}{{n}} \sum_{i=1}^n T_i
\end{equation}

\noindent
In cases where timing values are only available as a proportion of human-model response time similarity (e.g., $T_\mathcal{M} = 0.59=59\%$), this value is directly used:

\begin{equation}
T_\mathcal{M} = \text{empirical estimate} \in [0, 1]
\end{equation}
Although in cognitive modeling it is often expected that models that incorporate a richer set of constraints also provide a better fit to human data, it is important to note that this relation is not assumed here to hold \textit{a priori}. 
Within the present framework, indeed, Performance Match is not intended as a direct consequence of the Functional/Structural Ratio. A system may reproduce human-level outputs through mechanisms that differ from those posited by the target cognitive theory, just as a structurally constrained model may fail to achieve a strong Performance March when the selected constraints capture only a limited portion of the causal organization underlying the target phenomenon. For this reason, the two dimensions are kept distinct and considered independently.

\subsection{SME/CogSketch}

We assess the performance of both SME/CogSketch and LLMs on the standard version of Raven’s Progressive Matrices (RSPM) [\cite{raven1992manual}]. Introduced in [\cite{raven1938}], RPMs are widely accepted as a standard measure of intelligence in the broad sense. Here, more strictly, we consider them as a testbed for assessing abilities in the domain of relational (analogical) mapping. As shown by [\cite{SME-RPMs}], there is growing evidence that structural alignment processes performed in analogical reasoning are behind the visual comparison capabilities involved in solving tasks related to RPMs. Further observations on the test are required: classical psychological research from \cite{jensen1982reaction} and \cite{salthouse1993influence} has demonstrated significant performance differences between age groups, with peak scores observed in young adults (mean age = 20 years) and a noticeable decline with increasing age; also, the test itself is diversified into a variety of difficulty levels. Therefore, for simplicity, we will compare the models' results with the median performance (50th percentile) of the best-performing human group, as reported in \cite{raven1992manual}, Tab. V.

Model accuracy data in RSPM is available for SME/CogSketch in [\cite{SME-RPMs}]. SME/CogSketch were tested on sections B-C-D-E of the RSPM, achieving an accuracy score of 44/48, which was converted to 55/60 (91.66\%) on the overall test. These results far exceed the median performance of human subjects from the best-performing group (44/60, or 73.33\%), aligning instead with the median score of the 95th percentile. 
Regarding the analysis of error patterns, \cite{SME-RPMs} have demonstrated a similarity effect between SME/CogSketch's errors and those of human subjects: the four matrices that SME/CogSketch failed to solve correctly are among the six most difficult problems in tests conducted on human participants. According to the authors, the model is accurate in terms of performance-matching, as it replicates both the abilities and limitations of an above-average human. Consequently, its failures serve to explain and predict human performance.

\cite{lovett2009solving} combined SME and CogSketch to solve Geometric Analogy Problems as introduced by [\cite{evans1968program}], which involve visual transformations of simple geometric figures. The authors reconstructed the original 20 Evans problems in PowerPoint and gave them to both human participants (N=34) and SME/CogSketch. For 9 of the 20 problems, all participants chose the same answer, while over 90\% chose the same answer for 7 additional problems. On average, a mean human accuracy of 93\% across all 20 problems is estimated, defined the accuracy of a human participant as the probability of selecting the majority-chosen option. Time required to solve a problem varied from 4.5 to 26.7 s, with a mean of 9.6. SME/CogSketch chose the preferred human answer on all 20 Evans problems. As for error patterns, Lovett et al. (2009) report that on Problem 17, where the correctness of the answer depends on whether the subject privileges reflection or rotation, the degree of disagreement between participants (56\% selecting the preferred answer (Answer 4), 21\% selecting Answer 2 and 15\% selecting Answer 1), is reproduced by the model: it assigns the highest match score to Answer 4 (0.94), but also attributes substantial scores to the two remaining choices (0.72 for Answer 2 and 0.40 for Answer 1), confirming that it also mirrors the human error distribution. 
SME/CogSketch also demonstrated a good correspondence with human reaction time data. Specifically, the Normal mode showed a correlation of r = 0.59 with human reaction times, while the Shape-Identicality mode achieves r = 0.75. This indicates that the problems perceived as more difficult by participants (longer response times) also posed greater processing demands for the model.

\subsection{\textbf{MET}\textsuperscript{CL}}

In Section 3 of their paper, \cite{lieto2025delta} conduct a twofold evaluation of \textbf{MET}\textsuperscript{CL}. To assess its performance in metaphor classification and generation, they combined automatic benchmarking against existing systems with a human-centred assessment. 
In the automatic evaluation phase, \textbf{MET}\textsuperscript{CL} was tested using two distinct metaphor datasets: the NN-450 corpus (obtained by merging three existing datasets: the Gordon Metaphor corpus by [\cite{gordon2015corpus}], the VU Amsterdam Metaphor Corpus [\cite{krennmayr2017vu}, and the Metaphor Detection Dataset [\cite{mensa2018grasping}]), and the curated set of metaphorical examples from the MetaNet taxonomy [\cite{dodge2015metanet}]. The system was found to successfully generate metaphorical representations for the vast majority of the NN-450 items (98.44\%) and for over half of the MetaNet examples (55.23\%). These outputs were subsequently used to augment both LLMs and symbolic classifiers in zero-shot and few-shot learning scenarios. Across all settings, the inclusion of \textbf{MET}\textsuperscript{CL} outputs resulted in consistent improvements in both precision and recall. For instance, in the zero-shot classification setting, GPT-4o exhibited an increase in precision from 37.58\% to 55.33\% and in recall from 33.69\% to 49.94\%.

To complement this evaluation, a human evaluation analysis was conducted to assess the perceived plausibility of the metaphorical representations generated by \textbf{MET}\textsuperscript{CL}. 70 human judges (primarily researchers and students) were tasked with rating the quality of the compound concepts produced by the system. 630 metaphorical expressions (from NN-450 and MetaNet) were processed by \textbf{MET}\textsuperscript{CL}. Each output included the top-ranked typical properties (e.g., “infectious: 0.909”) of the generated metaphor. Participants rated each output on a scale from 1 to 10, evaluating how well the listed properties reflected the intended metaphorical meaning.

The results of this human evaluation indicate a moderate to strong degree of acceptance of \textbf{MET}\textsuperscript{CL}’s output. The overall mean rating across 630 evaluated metaphors was 5.99 out of 10 (median=6, SD=2.41). Given the recent development of the system, \textbf{MET}\textsuperscript{CL} has not yet been evaluated on standardized benchmarks that include human performance scores. Consequently, in order to estimate \textbf{MET}\textsuperscript{CL}’s accuracy in the metaphor generation task relative to human capabilities, we define the human baseline as the upper bound of the evaluation scale — that is, a score of 10 out of 10, representing the maximum degree of plausibility attributed to a generated metaphor by human judges (10/10 = 100\%). The accuracy delta is then computed as the deviation from this optimal score. In the case under analysis, the mean human rating is 5.99 out of 10, which corresponds to an approximate deviation of $\Delta_i=–40\%$. The current study does not allow for a direct comparison between \textbf{MET}\textsuperscript{CL}’s error patterns and those observed in human metaphor production or interpretation. Also, no response time data were collected or reported.

\subsection{LLMs}

As shown in the previous section, LLMs have been tested on a wide range of analogical tasks, such as: 1. Matrix reasoning problems (RPM presented in textual format), 2. Letter string analogies [\cite{hofstadter1994copycat}], 3. Four-term verbal analogies [UCLA VAT; \cite{sternberg1980developmental}, SAT analogy problems (\cite{turney2003combining}, \cite{jones2022differential})], 4. Story analogies.

Accuracy data on RSPM is available for Large Language Models in [\cite{RPM-LLMs}] (Mistral, Llama), and in [\cite{webb2023emergent}] (GPT-3 Text-Davinci-003): Mistral, Llama and GPT-3 scored 85\%, 88\% and 80\% respectively. Besides accuracy, a similarity in error patterns has been observed: \cite{RPM-LLMs} showed that performances by Mistral and Llama progressively declined as the complexity of the matrices increased, akin to human subjects and SME/CogSketch.

For letter-string analogy, \cite{lewis2024evaluating} replicated and extended the experimental protocol developed by \cite{webb2023emergent}, introducing a battery of counterfactual problems. These tasks included alphabetic sequences with permuted letters and non-letter symbol strings, designed to deviate from patterns likely encountered during pre-training. They evaluated three GPT models (GPT-3, GPT-3.5, and GPT-4) alongside a cohort of 136 human participants. The final results show that human performance is significantly higher compared to any of the language models. The average human accuracy was 75.3\% (95\% CI: [73.4\%, 77.3\%]), whereas the GPT models in the zero-shot experimental setting obtained considerably lower results: GPT-3: 48.8\% (95\% CI: [46.7\%, 50.9\%]); GPT-3.5: 35.0\% (95\% CI: [33.0\%, 37.0\%]); GPT-4: 45.2\% (95\% CI: [43.1\%, 47.3\%]). Notably, language models perform significantly worse when exposed to even small variations, like permuted alphabets.

Beyond accuracy scores, the authors conducted a qualitative error analysis to better understand the nature of incorrect responses across human and GPT model participants. Errors were manually categorised into four distinct types: 1. Alternative Rule. A plausible but unintended rule is applied, such as replacing the last letter with a fixed character rather than executing a genuine transformation; 2. Incorrect Rule Use. A transformation is attempted, but the inferred rule is inconsistent with the source-target pair; 3. Wrong. The answer is related to the target string but does not follow any coherent or inferable rule; 4. Completely Wrong. The answer is both unrelated and incoherent with respect to the target, often indicating random or nonsensical output.
Humans most frequently generated errors attributable to the \emph{Alternative Rule} category (38.59\%). GPT-4 exhibited similar behaviour to a lesser extent (22.00\%), while GPT-3 showed minimal use of alternative strategies (5.81\%). Both GPT-3 and GPT-4 committed a greater proportion of \emph{Incorrect Rule Use} and \emph{Wrong} errors. Notably, GPT-3’s largest error category was \emph{Wrong} (55.48\%).

\textsc{AnaloBench} [\cite{ye2024analobench}] was developed to test LLMs' analogical reasoning over natural language stories, instead of word-level analogies. The authors collected 340 pairs of analogous stories produced by human annotators, and tested models of varying sizes and architectures (GPT-4 [\cite{gpt4}], GPT-3.5 [\cite{brown2020}], LLaMA2-chat [Touvron et al., (2023)], XwinLM (Xwin-LM, 2023), WizardLM (Xu et al., 2023), Tulu2 (Ivison et al., 2023), Zephyr (Tunstall et al., 2023), Claude-v2 (Anthropic, 2023), etc.) on choosing the most fitting analogy from 4 options in 1-sentence, 10-sentence, and 30-sentence stories scenarios. While bigger models guaranteed better performances, the scaling effect decreased as the stories became longer. Tests were conducted both on a small subset and on the full story bank. On the small bank, human baseline scores were: 96\% (1-sentence), 72.5\% (10-sentence), and 73.3\% (30-sentence), while the SOTA model (GPT-4) scored 89.1\%, 66.5\%, and 60.7\% respectively ($\Delta_{i,j,k}= 6.9\%, 6\%, 12.6\%$). Scores declined consistently on the full bank, with GPT-4 and Claude approaching random level performances. Human baseline scores were only collected for the mini story bank due to the impractical cost of annotating the full set (although such annotation was technically feasible). As a result, we adopt the mini-bank scores as a reference. Readers should note that this is a charitable comparison: model performance on the full bank is considerably lower, suggesting the actual human-model gap is likely even larger, since there is no empirical reason to expect a drop in human accuracy when moving to the full dataset.

Additionally, although a slight accuracy drop in both humans and LLMs is registered as sentence length increases, the much more pronounced drop in LLMs (particularly GPT-4) suggests that error patterns differ. The lack of a comparable drop in human performance in longer contexts suggests that the difficulties encountered by LLMs might not be the same as those encountered by human annotators.

A variety of LLMs has been evaluated on \textsc{StoryAnalogy} [\cite{storyanalogy}], a benchmark comprising 24K story pairs across diverse domains with human annotations. The results show that both analogy identification and generation at the story level remain challenging, even for advanced LLMs. On multiple-choice tasks such as ``Which candidate story is the best creative analogy for the source story?'', ChatGPT and GPT-3 respectively achieved accuracies of 44.2\%, 34.1\%, and 33.1\% (0/1/3-shot), and 35.8\%, 29.2\%, and 32.3\%, while random choice yields 25\% and the human baseline reached 85.7\%. The SOTA model, Flan-T5-XXL [\cite{flan}], obtained 45.0\%, 46.3\%, and 45.0\% in the respective settings. For the generation task, based on human evaluations, ChatGPT was the top-performing model in the few-shot setting, achieving an average quality score of 82.7\%, with a deviation of -17.3\% from the human-perceived optimal.

\subsection{Results comparison for Performance Match}

\begin{table}[H]
    \centering
    \renewcommand{\arraystretch}{1.3}
    \footnotesize
    \begin{tabularx}{\textwidth}{ >{\centering\arraybackslash}X 
                                 >{\centering\arraybackslash}X 
                                 >{\centering\arraybackslash}X 
                                 >{\centering\arraybackslash}X 
                                 >{\centering\arraybackslash}X 
                                 >{\centering\arraybackslash}X 
                                 >{\centering\arraybackslash}X 
                                 >{\centering\arraybackslash}X
                                 >{\centering\arraybackslash}X }
        \toprule
        \textbf{Model} & 
        \textbf{Bench.} &
        \textbf{Human Baseline} & 
        \textbf{Accuracy} &
        \textbf{Delta} &
        \textbf{Error Patterns} &
        \textbf{Resp. Time} &
        $\boldsymbol{PM_\mathcal{M}}$ \\
        \midrule
        \rowcolor{gray!10}
        \rowcolor{gray!10}
        \makecell{CogSketch/\\SME} &
        \makecell{RSPM} & 
        \makecell{0.733} & 
        \makecell{0.916} & 
        \makecell{+0.183} & 
        \makecell{1} & 
        \makecell{N/A} & 
        \makecell{\textbf{0.923}} \\
        \midrule
        \rowcolor{gray!10}
        \makecell{\textbf{MET\textsuperscript{CL}}} & Hum. Eval. & 1.000 & 0.599 & -0.401 & N/A & N/A & \textbf{0.714} \\
        \midrule
        \addlinespace[0.55ex]
        \rowcolor{lightgray}
        \multicolumn{8}{c}{\textbf{LLMs}} \\
        \midrule
        LLama & RSPM & 0.733 & 0.880 & +0.150 & -1 & N/A & 0.436 \\
        GPT-3.5 & Lett. String & 0.753 & 0.488 & -0.265 & -1 & N/A & 0.395 \\  
        GPT-4o & ANALOB. (1-sent.) & 0.960 & 0.891 & -0.069 & -1 & N/A & 0.468 \\
        GPT-4o & ANALOB. (10-sent.) & 0.725 & 0.665 & -0.060 & -1 & N/A & 0.472 \\
        GPT-4o & ANALOB. (30-sent.) & 0.733 & 0.607 & -0.126 & -1 & N/A & 0.444 \\
        Flan-T5 & S.AN. T2 (0-shot) & 0.857 & 0.450 & -0.407 & N/A & N/A & 0.711 \\
        Flan-T5 & S.AN. T2 (1-shot) & 0.857 & 0.463 & -0.394 & N/A & N/A & 0.717 \\
        Flan-T5 & S.AN. T2 (3-shot) & 0.857 & 0.450 & -0.407 & N/A & N/A & 0.711 \\
        ChatGPT & S.AN. T1 (Hum. Eval.) & 1.000 & 0.827 & -0.173 & N/A & N/A & 0.853 \\
        \rowcolor{gray!10}
\textbf{LLMs (avg)} & \textbf{--} & \textbf{0.830} & \textbf{0.636} & \textbf{-0.228} & \textbf{--} & \textbf{--} & \textbf{0.578} \\

        \bottomrule
    \end{tabularx}

    \captionsetup{width=0.9\textwidth}
    \caption{Model performance comparison against human baselines, with accuracy and relative delta reported as decimal values. RSPM = Raven’s Standard Progressive Matrices; Hum. Eval. = Human Evaluation; Lett. String = Letter String Analogies; ANALOB. = \textsc{AnaloBench} (1/10/30 sentence); S.AN. T2 = \textsc{StoryAnalogy} Task 2 (Multiple Choice 0/1/3-shot); S.AN. T1 = \textsc{StoryAnalogy} Task 1 (Generation).}
\end{table}

\section{Unified Metric for Cognitive Plausibility}

To evaluate the overall cognitive plausibility of a model, we define a composite metric ($\mathcal{CP}_\mathcal{M}$) which integrates the three distinct dimensions analyzed through the Minimal Cognitive Grid: \textit{Functional/Structural Ratio}, \textit{Generality}, and \textit{Performance Match}. The final score is computed as the  weighted sum:

\begin{equation}
\mathcal{CP}_\mathcal{M} = \lambda \cdot  FSR'_\mathcal{M} +  \mu \cdot G_\mathcal{M} + \nu \cdot PM_\mathcal{M} \quad \text{with} \quad \lambda + \mu + \nu = 1
\end{equation}

\noindent
With $G_\mathcal{M}$ being the Generality score, and $PM_\mathcal{M}$ the Performance Match score, while $\lambda, \mu, \nu$ are non-negative weights that sum to 1 and reflect the relative importance of each dimension. 
In the present work, we assign a higher weight to the Functional/Structural Ratio, since it captures the degree to which a model’s mechanisms align with the structural principles of human analogical reasoning. Consequently, we set the weights as follows:

\begin{equation}
(\lambda, \mu, \nu)= (0.5, 0.25, 0.25)
\end{equation}

\noindent So that $FSR'_\mathcal{M}$ contributes as much as the other two combined. This weighting scheme reflects the primary aim of the framework, namely, to assess whether a model can serve as a testbed for theorizing about psychological phenomena in computational cognitive science. If this is the aim, then adherence to the mechanisms and constraints that are taken to underlie the target cognitive phenomenon must be given priority. As argued by [\cite{kaplancraver2011, harbecke2021}], the explanatory relevance of a model requires a mapping between the elements of the model and the elements of the mechanism responsible for the explanandum. In this sense, the functional/structural ratio is the dimension that anchors our evaluation to the explanatory profile of the model, rather than one desideratum among others.
This does not imply that generality and performance match are of secondary importance. On the contrary, both are essential, and generality in particular is a major desideratum for any candidate cognitive model. For this reason, one could adopt alternative non-equal weighting schemes (for example, $\lambda = 0.5$, $\mu = 0.3$, and $\nu = 0.2$). The crucial point, however, is that they should not compensate for the absence of the relevant mechanisms. A model may generalize broadly and fit behavioral data closely while still remaining weak as a cognitive explanation if its success is not grounded in the kind of internal organization that is relevant to the phenomenon at issue [\cite{pittmyung2002}]. 
At the same time, we recognize that alternative weighting schemes may legitimately reflect different theoretical priorities. To make this explicit, Table~5 also includes the aggregate scores obtained under equal weighting, allowing readers to inspect how the comparative picture changes under a different weighting assumption.

\subsection{Computing the Overall Cognitive Plausibility Score with the MCG}

The following table reports the scores for each model across the three  sub-metrics of our framework: Functional/Structural Ratio ($FSR'_\mathcal{M}$), Generality ($G_\mathcal{M}$), and Performance Match ($PM_\mathcal{M}$). The two gray blocks report the overall cognitive plausibility score $\mathcal{CP}_{\mathcal M}$ under two alternative weighting schemes: (1) Non-equal, that assigns greater weight to $FSR'_{\mathcal M}$ ($0.5, 0.25, 0.25$); (2) Equal, that assigns the same weight to each dimension ($1/3$). For each weighting scheme, the $\mathcal{CP}_{\mathcal M}$ scores are reported both with standard generality ($G_\mathcal{M}$) and with the variant $G_{\mathcal M}(1)$ introduced in Section 3 (Tab. \ref{tab:task_domains}).

\begin{table}[H]
\centering
\renewcommand{\arraystretch}{1.15}
\setlength{\tabcolsep}{4pt}
\footnotesize

\resizebox{\linewidth}{!}{%
\begin{tabular}{
l c c c c
>{\columncolor{gray!15}}c >{\columncolor{gray!15}}c
>{\columncolor{gray!25}}c >{\columncolor{gray!25}}c
}
\toprule
& \multicolumn{4}{c}{\textbf{Base metrics}}
& \multicolumn{2}{c}{\cellcolor{gray!15}\textbf{Non-equal weighting}}
& \multicolumn{2}{c}{\cellcolor{gray!25}\textbf{Equal weighting}} \\
\cmidrule(lr){2-5} \cmidrule(lr){6-7} \cmidrule(lr){8-9}

\textbf{Model}
& $\boldsymbol{FSR'_\mathcal{M}}$
& $\boldsymbol{G_\mathcal{M}}$
& $\boldsymbol{G_\mathcal{M}(1)}$
& $\boldsymbol{PM_\mathcal{M}}$
& \cellcolor{gray!15}\textbf{$\mathcal{CP}_\mathcal{M}$}
& \cellcolor{gray!15}\textbf{$\mathcal{CP}_\mathcal{M}(1)$}
& \cellcolor{gray!25}\textbf{$\mathcal{CP}_\mathcal{M}$}
& \cellcolor{gray!25}\textbf{$\mathcal{CP}_\mathcal{M}(1)$} \\
\midrule

CogSketch
& \textbf{0.606} & 0.125 & 0.200
& \textbf{0.923} & \textbf{0.565} & \textbf{0.584} & \textbf{0.551} & \textbf{0.576} \\

SME
& \textbf{0.606} & 0.063 & 0.100
& \textbf{0.923} & 0.549 & 0.559 & 0.531 & 0.543 \\

\textbf{MET\textsuperscript{CL}}
& 0.407 & 0.125 & 0.200
& 0.714 & 0.413 & 0.432 & 0.415 & 0.440 \\

LLMs (avg)
& 0.109 & \textbf{0.500} & \textbf{0.800}
& 0.578 & 0.324 & 0.399 & 0.396 & 0.496 \\

\bottomrule
\end{tabular}%
}
\captionsetup{width=0.9\textwidth}
\caption{Comparison of the models across the three main dimensions of the framework: functional-structural ratio ($FSR'_{\mathcal M}$), generality ($G_{\mathcal M}$ and its variant $G_{\mathcal M}(1)$), and performance match ($PM_{\mathcal M}$). The two gray blocks report the overall cognitive plausibility score $\mathcal{CP}_{\mathcal M}$ under two alternative weighting schemes. In the non-equal weighting scheme, greater importance is assigned to the functional-structural ratio ($0.5, 0.25, 0.25$), whereas in the equal weighting scheme all three dimensions are weighted uniformly ($1/3$ each). For both schemes, the overall score is reported once using standard generality and once using the alternative generality variant $G_{\mathcal M}(1)$.}
\label{tab:overall}
\end{table}

\noindent
SME and CogSketch obtain the highest scores on both $FSR'_{\mathcal M}$ (0.606) and $PM_{\mathcal M}$ (0.923) although with the limit of having been tested against a single benchmark. Their weakness lies instead in generality. Under the MCG evaluation, these systems appear cognitively grounded and aligned with human behavioral data, but their applicability remains limited to a very narrow range of tasks.
By contrast, LLMs obtain by far the highest generality scores ($G_{\mathcal M} = 0.500$ and $G_{\mathcal M}(1) = 0.800$), while scoring much lower on the functional-structural ratio (0.109), thus appearing general but weak with respect to their grounding in cognitive structures and mechanisms. 
\textbf{MET}\textsuperscript{CL} occupies an intermediate position. Its $FSR'_{\mathcal M}$ score (0.407) is higher than that of LLMs, which is consistent with the fact that it explicitly implements a heuristic grounded in the Categorization Theory of Metaphor. At the same time, its generality remains limited (0.125 / 0.200), and its performance match score (0.714) remains below that of SME and CogSketch.
Under the non-equal weighting scheme CogSketch obtains the highest overall score, followed by SME. \textbf{MET}\textsuperscript{CL} remains above LLMs in this setting,  despite its lower generality, reflecting its balance between cognitive inspiration and alignment with human behavioral data.
A small differentiation emerges under the equal weighting scheme: when all the dimensions are treated as equally important, the advantage of LLMs in generality becomes strong enough to compensate for their low $FSR'_{\mathcal M}$ score, and the overall cognitive plausibility score for LLMs surpasses that of \textbf{MET}\textsuperscript{CL} (the ranking remains unaltered for SME and CogSketch). This is a particularly enlightening result: LLMs, which are not built, by design, around any cognitively or biologically inspired constraint (on the epistemological differences between LLMs and human cognition, see \cite{quattrociocchi2025epistemological}), end up being ranked as more cognitively plausible than a model that explicitly implements a cognitively grounded heuristic. This paradoxical outcome suggests that a flat weighting scheme may be less appropriate for capturing the purpose of cognitive plausibility assessment within the MCG framework. On the other hand, it highlights the robusteness of the MCG and the limited sensitivity of its final ranking to the weighting assumptions being made, since this paradoxical outcome is the only result changing.  
Overall, in fact, the original weighting scheme of the Minimal Cognitive Grid (i.e., the one represented as Base Metrics in the table above) appears to be more compliant with the findings and consensus coming from the neuroscientific and psychological literature. In particular, assigning a major weight to sensorimotor tasks in the Generality rating is aligned with findings indicating that, in biological entities, these classes of tasks are fundamental for the development of high level cognitive skills. In addition, concerning the major importance assigned to the Functional/Structural ratio among the dimensions of the MCG, it is worth recalling that the original goal of the framework is to provide a quantitative projection of the explanatory power of artificial systems with respect to the biological systems taken as a source of inspiration for a particular task (or an ensemble of tasks) at hand. In other words, the main goal is to detect to what extent the computational approach used to design an AI system enables to provide not only the achievement of human-level (or super-human) results, but also the interpretation of such performances as the output of human-like computations. In doing so, it is evident that the mere comparison of behavioral performances represents an important and necessary component for such an assessment, but not the major one. A major role, in this respect, is played by the way such AI systems have been explicitly designed and by the biological/cognitive constraints they explicitly incorporate to carry out human-like computation. The functional/structural dimension is, in the MCG, the only one that can quantitatively assess such an issue by avoiding the trap of a renewed behaviorism [\cite{li2026ghost}]. Furthermore, the increasing acceptance of the original configuration of the Minimal Cognitive Grid across different fields seems to be confirmed by its growing adoption in neuroscience [\cite{milkowski2026confirmation}], bionic systems [\cite{lieto2022analyzing}], and artificial creativity [\cite{da2025artificial}].

\section{Conclusions and Future Work}

This study presents a formal operationalization of the Minimal Cognitive Grid (MCG), with an application to the comparative evaluation of computational models of analogy and metaphor. The framework combines three complementary dimensions — \emph{Functional/Structural Ratio}, \emph{Generality}, and \emph{Performance Match} — into a unified index of cognitive plausibility ($\mathcal{CP}_\mathcal{M}$). $FSR’_\mathcal{M}$ captures the degree of structural fidelity to constraints (processing mechanisms, heuristics, principles) derived from the Structure-Mapping and Metaphor-as-Categorization theories; Generality measures the breadth of functions that the systems are able to replicate across cognitive and sensory/motor abilities, grounded in the Cattell-Horn-Carroll taxonomy; Performance Match quantifies the empirical correspondence between model and human behavior across accuracy, error patterns, and response times. These components are then combined into a general index of cognitive plausibility ($\mathcal{CP}_\mathcal{M}$).

Empirical results show that SME and CogSketch achieve the highest structural scores ($FSR' = 0.606$), reflecting their explicit implementation of cognitive constraints such as one-to-one mapping, parallel connectivity, and systematicity. MET$^{\text{CL}}$ follows ($FSR' = 0.407$), with its satisfaction of the Categorization and Property Selection mechanisms, while LLMs exhibit very low structural fidelity ($FSR' = 0.109$). With respect to Generality, LLMs obtain the highest score ($G_{\mathcal M} = 0.500$ and $G_{\mathcal M}(1) = 0.800$) thanks to their wide functional range and multimodal versatility, whereas CogSketch ($0.125, 0.200$), MET$^{\text{CL}}$ ($0.125, 0.200$), and SME ($0.063, 0.100$) show more specialized profiles. In terms of behavioral alignment, SME and CogSketch once again outperform the others, showing high correspondence with human accuracy and error patterns on Raven’s Progressive Matrices, while MET$^{\text{CL}}$ and the LLMs reach intermediate scores. It must be acknowledged that, whereas LLMs were evaluated across multiple benchmarks, SME and CogSketch results derive from a single available dataset; their average performance might thus appear inflated compared to what would emerge from a broader evaluation.

The quantitative Minimal Cognitive Grid provides a theory-and model-agnostic evaluation framework for assessing how artificial systems align with the key dimensions of cognitive modeling through a unified, quantitative metric. Rather than endorsing or rejecting any specific paradigm, the framework establishes a common measurement language that can be applied across architectures.

From this perspective, the hitherto included empirical results serve as a validation of the framework: as expected, the scores reflect that LLMs display remarkable generality and behavioral performance, but they do not qualify as models of cognition in a mechanistic sense, as they lack explicit representational and process-level alignment with human reasoning. Classical structure-mapping systems, on the other hand, despite their limited generality, remain more faithful to such mechanisms. The MCG can make such comparisons explicit, transparent, reproducible, and offer a scalable methodology for the systematic assessment of cognitive plausibility across current and future generations of AI systems.

Some limitations affect the work and indicate directions for future refinement. 
First, a complete lack of execution time data prevented the inclusion of temporal metrics in the Performance Match component. Since human execution time is an indirect indicator of underlying processing strategies, this omission flattens the importance of $PM_\mathcal{M}$ to static, behavioral performance measures of cognition. Despite such lack of data, however, the multidimensional and prismatic assessment provided by the MCG allowed us to showcase robust outcomes also in the context of sensitivity analyses with diverse weighted schemes.
Although in lower proportions, data on error patterns remain sparse or inconsistent as well.
Third, the availability of benchmark data is uneven across model classes: LLMs benefit from a rich ecosystem of analogy and metaphor benchmarks, while SME, CogSketch, and MET$^{\text{CL}}$ are evaluated on far fewer datasets. 
This imbalance makes the LLMs’ performance estimates more stable than the ones of the other systems.

Future work should address these limitations by including (and also developing) richer datasets encompassing item-level human error distributions and response times.

Also, at the current stage, the weighting scheme for the structural constraints in Tab.\ref{tab:FSR}, despite being grounded in empirical analysis; is not automatically derived through data-driven methods. However, establishing this type of empirical grounding is not trivial and remains an open challenge and a future direction to take in order to further extend the generalizability of the MCG. Expanding the set of the evaluated models could further test the scalability of the methodology and clarify the boundaries of its applicability.

A second research trajectory concerns the transition from evaluation to explanation. 
By applying the MCG to systems with well-understood internal mechanisms, it could become possible to define reference ranges and thresholds for the three dimensions and for the global plausibility index, which would allow researchers to interpret a model’s cognitive status at a glance, determining the kinds of scientific, biological, or psychological hypotheses that it can plausibly support as a tool for testing and discovery. 
Addressing the methodological and empirical limitations discussed here would enable the quantitative MCG to evolve into a comprehensive evaluation standard, capable of mapping the landscape of cognitively inspired Artificial Intelligence.

\pagebreak


\bibliographystyle{apalike}
\bibliography{Bibliografia.bib}

\begin{thebibliography}{}

\bibitem[Achiam et~al., 2023]{gpt4}
Achiam, J., Adler, S., Agarwal, S., Ahmad, L., Akkaya, I., Aleman, F.~L.,
  Almeida, D., Altenschmidt, J., Altman, S., Anadkat, S., et~al. (2023).
\newblock Gpt-4 technical report.
\newblock {\em arXiv preprint arXiv:2303.08774}.

\bibitem[Barsalou, 2008]{barsalou2008grounded}
Barsalou, L.~W. (2008).
\newblock Grounded cognition.
\newblock {\em Annual Review of Psychology}, 59:617--645.

\bibitem[Bhavya et~al., 2022]{analogygenLLM}
Bhavya, B., Xiong, J., and Zhai, C. (2022).
\newblock Analogy generation by prompting large language models: A case study
  of {I}nstruct{GPT}.
\newblock In Shaikh, S., Ferreira, T., and Stent, A., editors, {\em Proceedings
  of the 15th International Conference on Natural Language Generation}, pages
  298--312, Waterville, Maine, USA and virtual meeting. Association for
  Computational Linguistics.

\bibitem[Bringsjord, 2011]{bringsjord2011psychometric}
Bringsjord, S. (2011).
\newblock Psychometric artificial intelligence.
\newblock {\em Journal of Experimental \& Theoretical Artificial Intelligence},
  23(3):271--277.

\bibitem[Brown et~al., 2020]{brown2020}
Brown et~al. (2020).
\newblock Language models are few-shot learners.

\bibitem[Carroll, 1993]{carroll1993human}
Carroll, J.~B. (1993).
\newblock {\em Human cognitive abilities: A survey of factor-analytic studies}.
\newblock Cambridge University Press.

\bibitem[Chollet et~al., 2025]{chollet2025arc}
Chollet, F., Knoop, M., Kamradt, G., Landers, B., and Pinkard, H. (2025).
\newblock Arc-agi-2: A new challenge for frontier ai reasoning systems.
\newblock {\em arXiv preprint arXiv:2505.11831}.

\bibitem[Chomsky, 2011]{chomsky2011language}
Chomsky, N. (2011).
\newblock Language and other cognitive systems. what is special about language?
\newblock {\em Language learning and development}, 7(4):263--278.

\bibitem[Chung et~al., 2024]{flan}
Chung, H.~W., Hou, L., Longpre, S., Zoph, B., Tay, Y., Fedus, W., Li, Y., Wang,
  X., Dehghani, M., Brahma, S., et~al. (2024).
\newblock Scaling instruction-finetuned language models.
\newblock {\em Journal of Machine Learning Research}, 25(70):1--53.

\bibitem[Clark, 1999]{clark1999embodiment}
Clark, A. (1999).
\newblock An embodied cognitive science?
\newblock {\em Trends in Cognitive Sciences}.

\bibitem[Clement and Gentner, 1991]{clement1991systematicity}
Clement, C.~A. and Gentner, D. (1991).
\newblock Systematicity as a selection constraint in analogical mapping.
\newblock {\em Cognitive science}, 15(1):89--132.

\bibitem[Cordeschi, 2002]{cordeschi2002discovery}
Cordeschi, R. (2002).
\newblock {\em The discovery of the artificial: Behavior, mind and machines
  before and beyond cybernetics}, volume~28.
\newblock Springer Science \& Business Media.

\bibitem[Crick, 1989]{crick1989recent}
Crick, F. (1989).
\newblock The recent excitement about neural networks.
\newblock {\em Nature}, 337(6203):129--132.

\bibitem[Da~Pelo, 2025]{da2025artificial}
Da~Pelo, M. (2025).
\newblock Artificial creativity: can there be creativity without cognition?
\newblock {\em AI \& SOCIETY}, pages 1--14.

\bibitem[Ding et~al., 2023]{ding2023fluid}
Ding, Z., Srinivasan, A., MacNeil, S., and Chan, J. (2023).
\newblock Fluid transformers and creative analogies: Exploring large language
  models’ capacity for augmenting cross-domain analogical creativity.
\newblock In {\em Proceedings of the 15th Conference on Creativity and
  Cognition}, pages 489--505.

\bibitem[Dodge et~al., 2015]{dodge2015metanet}
Dodge, E.~K., Hong, J., and Stickles, E. (2015).
\newblock Metanet: Deep semantic automatic metaphor analysis.
\newblock In {\em Proceedings of the Third Workshop on Metaphor in NLP}, pages
  40--49.

\bibitem[Evans, 1968]{evans1968program}
Evans, T. (1968).
\newblock A program for the solution of geometric-analogy intelligence test
  questions. semantic information processing, m. minsky.

\bibitem[Falkenhainer et~al., 1989a]{falkenhainer1989structure}
Falkenhainer, B., Forbus, K.~D., and Gentner, D. (1989a).
\newblock The structure-mapping engine: Algorithm and examples.
\newblock {\em Artificial intelligence}, 41(1):1--63.

\bibitem[Falkenhainer et~al., 1989b]{falkenhainerSME}
Falkenhainer, B., Forbus, K.~D., and Gentner, D. (1989b).
\newblock The structure-mapping engine: Algorithm and examples.
\newblock {\em Artificial intelligence}, 41(1):1--63.

\bibitem[Forbus et~al., 2011]{cogsketch}
Forbus, K. et~al. (2011).
\newblock Cogsketch: Sketch understanding for cognitive science research and
  for education.
\newblock {\em Topics in Cognitive Science}, 3(4):648--666.

\bibitem[Forbus et~al., 2008]{forbus2008cogsketch}
Forbus, K., Lovett, A., Lockwood, K., Wetzel, J., Matuk, C., Jee, B., and
  Usher, J. (2008).
\newblock Cogsketch.
\newblock In {\em Proceedings of the 23rd national conference on Artificial
  intelligence-Volume 3}, pages 1878--1879.

\bibitem[Forbus et~al., 2005]{forbus2005structure}
Forbus, K.~D., Lovett, A., Tomai, E., and Usher, J. (2005).
\newblock A structure mapping model for solving geometric analogy problems.
\newblock In {\em Proceedings of the Annual Meeting of the Cognitive Science
  Society}, volume~27.

\bibitem[Fu et~al., 2024]{fu2024mme}
Fu, C., Zhang, Y.-F., Yin, S., Li, B., Fang, X., Zhao, S., Duan, H., Sun, X.,
  Liu, Z., Wang, L., et~al. (2024).
\newblock Mme-survey: A comprehensive survey on evaluation of multimodal llms.
\newblock {\em arXiv preprint arXiv:2411.15296}.

\bibitem[Gallese et~al., 1996]{gallese1996action}
Gallese, V. et~al. (1996).
\newblock Action recognition in the premotor cortex.
\newblock {\em Brain}.

\bibitem[Gentner, 1983]{gentnerstructuremapping}
Gentner, D. (1983).
\newblock Structure-mapping: A theoretical framework for analogy.
\newblock {\em Cognitive science}, 7(2):155--170.

\bibitem[Gentner, 1988]{gentner1988metaphor}
Gentner, D. (1988).
\newblock Metaphor as structure mapping: The relational shift.
\newblock {\em Child development}, pages 47--59.

\bibitem[Gentner and Bowdle, 2008]{gentner2008metaphor}
Gentner, D. and Bowdle, B. (2008).
\newblock Metaphor as structure-mapping.
\newblock {\em The Cambridge handbook of metaphor and thought}, 109:128.

\bibitem[Gentner et~al., 2001]{gentner2001analogical}
Gentner, D., Holyoak, K.~J., and Kokinov, B.~N. (2001).
\newblock {\em The analogical mind: Perspectives from cognitive science}.
\newblock MIT press.

\bibitem[Gentner and Jameson, 2006]{gentner2006systematicity}
Gentner, D. and Jameson, J. (2006).
\newblock Systematicity as a processing constraint on feature centrality.
\newblock In {\em Proceedings of the Annual Meeting of the Cognitive Science
  Society}, volume~28.

\bibitem[Gentner and Markman, 1997]{gentnermarkman}
Gentner, D. and Markman, A.~B. (1997).
\newblock Structure mapping in analogy and similarity.
\newblock {\em American psychologist}, 52(1):45.

\bibitem[Gentner and Toupin, 1986]{gentner1986systematicity}
Gentner, D. and Toupin, C. (1986).
\newblock Systematicity and surface similarity in the development of analogy.
\newblock {\em Cognitive science}, 10(3):277--300.

\bibitem[Glucksberg and McGlone, 2001]{glucksberg2001understanding}
Glucksberg, S. and McGlone, M.~S. (2001).
\newblock {\em Understanding figurative language: From metaphor to idioms}.
\newblock Number~36. Oxford University Press.

\bibitem[Gordon et~al., 2015]{gordon2015corpus}
Gordon, J., Hobbs, J.~R., May, J., Mohler, M., Morbini, F., Rink, B.,
  Tomlinson, M., and Wertheim, S. (2015).
\newblock A corpus of rich metaphor annotation.
\newblock In {\em Proceedings of the Third Workshop on Metaphor in NLP}, pages
  56--66.

\bibitem[Hampton, 1987]{hampton1987inheritance}
Hampton, J.~A. (1987).
\newblock Inheritance of attributes in natural concept conjunctions.
\newblock {\em Memory \& Cognition}, 15(1):55--71.

\bibitem[Harbecke, 2021]{harbecke2021}
Harbecke, J. (2021).
\newblock The methodological role of mechanistic-computational models in
  cognitive science.
\newblock {\em Synthese}, 199:12785--12807.

\bibitem[Hendrycks et~al., 2021]{hendrycks2021measuring}
Hendrycks, D., Burns, C., Kadavath, S., Arora, A., Basart, S., Tang, E., Song,
  D., and Steinhardt, J. (2021).
\newblock Measuring mathematical problem solving with the math dataset.
\newblock {\em arXiv preprint arXiv:2103.03874}.

\bibitem[Hodel and West, 2023]{hodel2023response}
Hodel, D. and West, J. (2023).
\newblock Response: Emergent analogical reasoning in large language models.
\newblock {\em arXiv preprint arXiv:2308.16118}.

\bibitem[Hofstadter and Mitchell, 1995]{hofstadter1994copycat}
Hofstadter, D.~R. and Mitchell, M. (1995).
\newblock The copycat project: A model of mental fluidity and analogy-making.
\newblock {\em Advances in connectionist and neural computation theory},
  2(205-267):2--3.

\bibitem[Holyoak, 2012]{holyoak2012analogy}
Holyoak, K.~J. (2012).
\newblock Analogy and relational reasoning.
\newblock {\em The Oxford handbook of thinking and reasoning}, pages 234--259.

\bibitem[Holyoak and Stamenkovi{\'c}, 2018]{holyoak2018metaphor}
Holyoak, K.~J. and Stamenkovi{\'c}, D. (2018).
\newblock Metaphor comprehension: A critical review of theories and evidence.
\newblock {\em Psychological bulletin}, 144(6):641.

\bibitem[Jensen, 1982]{jensen1982reaction}
Jensen, A.~R. (1982).
\newblock Reaction time and psychometric g.
\newblock In {\em A model for intelligence}, pages 93--132. Springer.

\bibitem[Jiayang et~al., 2023]{storyanalogy}
Jiayang, C., Qiu, L., Chan, T.~H., Fang, T., Wang, W., Chan, C., Ru, D., Guo,
  Q., Zhang, H., Song, Y., et~al. (2023).
\newblock Storyanalogy: Deriving story-level analogies from large language
  models to unlock analogical understanding.
\newblock {\em arXiv preprint arXiv:2310.12874}.

\bibitem[Jones et~al., 2022]{jones2022differential}
Jones, L.~L., Kmiecik, M.~J., Irwin, J.~L., and Morrison, R.~G. (2022).
\newblock Differential effects of semantic distance, distractor salience, and
  relations in verbal analogy.
\newblock {\em Psychonomic bulletin \& review}, 29(4):1480--1491.

\bibitem[Kaplan and Craver, 2011]{kaplancraver2011}
Kaplan, D.~M. and Craver, C.~F. (2011).
\newblock The explanatory force of dynamical and mathematical models in
  neuroscience: A mechanistic perspective.
\newblock {\em Philosophy of Science}, 78(4):601--627.

\bibitem[Kim et~al., 2023]{kim2023metaphorian}
Kim, J., Suh, S., Chilton, L.~B., and Xia, H. (2023).
\newblock Metaphorian: Leveraging large language models to support extended
  metaphor creation for science writing.
\newblock In {\em Proceedings of the 2023 ACM Designing Interactive Systems
  Conference}, pages 115--135.

\bibitem[Kintsch, 1994]{kintsch1994text}
Kintsch, W. (1994).
\newblock Text comprehension, memory, and learning.
\newblock {\em American psychologist}, 49(4):294.

\bibitem[Klenk et~al., 2005]{klenk2005solving}
Klenk, M., Forbus, K.~D., Tomai, E., Kim, H., and Kyckelhahn, B. (2005).
\newblock Solving everyday physical reasoning problems by analogy using
  sketches.
\newblock In {\em PROCEEDINGS OF THE NATIONAL CONFERENCE ON ARTIFICIAL
  INTELLIGENCE}, volume~20, page 209. Menlo Park, CA; Cambridge, MA; London;
  AAAI Press; MIT Press; 1999.

\bibitem[Kojima et~al., 2022]{kojima2022large}
Kojima, T., Gu, S.~S., Reid, M., Matsuo, Y., and Iwasawa, Y. (2022).
\newblock Large language models are zero-shot reasoners.
\newblock {\em Advances in neural information processing systems},
  35:22199--22213.

\bibitem[Krennmayr and Steen, 2017]{krennmayr2017vu}
Krennmayr, T. and Steen, G. (2017).
\newblock Vu amsterdam metaphor corpus.
\newblock In {\em Handbook of linguistic annotation}, pages 1053--1071.
  Springer.

\bibitem[Lee et~al., 2025]{ARC-LLMs}
Lee, S., Sim, W., Shin, D., Seo, W., Park, J., Lee, S., Hwang, S., Kim, S., and
  Kim, S. (2025).
\newblock Reasoning abilities of large language models: In-depth analysis on
  the abstraction and reasoning corpus.
\newblock {\em ACM Trans. Intell. Syst. Technol.}
\newblock Just Accepted.

\bibitem[Levesque et~al., 2012]{winograd}
Levesque, H., Davis, E., and Morgenstern, L. (2012).
\newblock The winograd schema challenge.
\newblock In {\em Thirteenth international conference on the principles of
  knowledge representation and reasoning}.

\bibitem[Lewis and Mitchell, 2024]{lewis2024evaluating}
Lewis, M. and Mitchell, M. (2024).
\newblock Evaluating the robustness of analogical reasoning in large language
  models.
\newblock {\em arXiv preprint arXiv:2411.14215}.

\bibitem[Li et~al., 2023a]{li2023seed}
Li, B., Wang, R., Wang, G., Ge, Y., Ge, Y., and Shan, Y. (2023a).
\newblock Seed-bench: Benchmarking multimodal llms with generative
  comprehension.
\newblock {\em arXiv preprint arXiv:2307.16125}.

\bibitem[Li et~al., 2023b]{li2023evaluating}
Li, Y., Du, Y., Zhou, K., Wang, J., Zhao, W.~X., and Wen, J.-R. (2023b).
\newblock Evaluating object hallucination in large vision-language models.
\newblock {\em arXiv preprint arXiv:2305.10355}.

\bibitem[Li et~al., 2026]{li2026ghost}
Li, Z., Wang, Y., and Wu, Q. (2026).
\newblock The ghost of behaviorism: critical reflections on methodological
  limitations in the research of large language models psychology.
\newblock {\em Cognitive Systems Research}, page 101445.

\bibitem[Lieder and Griffiths, 2020]{lieder2020resource}
Lieder, F. and Griffiths, T.~L. (2020).
\newblock Resource-rational analysis: Understanding human cognition as the
  optimal use of limited computational resources.
\newblock {\em Behavioral and brain sciences}, 43:e1.

\bibitem[Lieto, 2021]{lieto2021cognitive}
Lieto, A. (2021).
\newblock {\em Cognitive design for artificial minds}.
\newblock Routledge.

\bibitem[Lieto, 2022]{lieto2022analyzing}
Lieto, A. (2022).
\newblock Analyzing the explanatory power of bionic systems with the minimal
  cognitive grid.
\newblock {\em Frontiers in Robotics and AI}, 9:888199.

\bibitem[Lieto and Pozzato, 2018]{lieto2018description}
Lieto, A. and Pozzato, G.~L. (2018).
\newblock A description logic of typicality for conceptual combination.
\newblock In {\em International symposium on methodologies for intelligent
  systems}, pages 189--199. Springer.

\bibitem[Lieto and Pozzato, 2020]{lieto2020description}
Lieto, A. and Pozzato, G.~L. (2020).
\newblock A description logic framework for commonsense conceptual combination
  integrating typicality, probabilities and cognitive heuristics.
\newblock {\em Journal of Experimental \& Theoretical Artificial Intelligence},
  32(5):769--804.

\bibitem[Lieto et~al., 2025]{lieto2025delta}
Lieto, A., Pozzato, G.~L., and Zoia, S. (2025).
\newblock The delta of thought: Channeling rivers of commonsense knowledge in
  the sea of metaphorical interpretations.
\newblock In Kwok, J., editor, {\em Proceedings of the Thirty-Fourth
  International Joint Conference on Artificial Intelligence, {IJCAI-25}}, pages
  10316--10324. International Joint Conferences on Artificial Intelligence
  Organization.
\newblock Human-Centred AI.

\bibitem[Liu et~al., 2024]{mmbench}
Liu, Y., Duan, H., Zhang, Y., Li, B., Zhang, S., Zhao, W., Yuan, Y., Wang, J.,
  He, C., Liu, Z., Chen, K., and Lin, D. (2024).
\newblock Mmbench: Is your multi-modal model an all-around player?

\bibitem[Loru et~al., 2025]{loru2025simulation}
Loru, E., Nudo, J., Di~Marco, N., Santirocchi, A., Atzeni, R., Cinelli, M.,
  Cestari, V., Rossi-Arnaud, C., and Quattrociocchi, W. (2025).
\newblock The simulation of judgment in llms.
\newblock {\em Proceedings of the National Academy of Sciences},
  122(42):e2518443122.

\bibitem[Lovett and Forbus, 2017]{lovett2017modeling}
Lovett, A. and Forbus, K. (2017).
\newblock Modeling visual problem solving as analogical reasoning.
\newblock {\em Psychological review}, 124(1):60.

\bibitem[Lovett et~al., 2007]{lovett2007analogy}
Lovett, A., Forbus, K., and Usher, J. (2007).
\newblock Analogy with qualitative spatial representations can simulate solving
  raven's progressive matrices.
\newblock In {\em Proceedings of the Annual Meeting of the Cognitive Science
  Society}, volume~29.

\bibitem[Lovett et~al., 2010]{SME-RPMs}
Lovett, A., Forbus, K., and Usher, J. (2010).
\newblock A structure-mapping model of raven's progressive matrices.
\newblock In {\em Proceedings of the Annual Meeting of the Cognitive Science
  Society}, volume~32.

\bibitem[Lovett et~al., 2009a]{lovett2009using}
Lovett, A., Gentner, D., Forbus, K., and Sagi, E. (2009a).
\newblock Using analogical mapping to simulate time-course phenomena in
  perceptual similarity.
\newblock {\em Cognitive Systems Research}, 10(3):216--228.

\bibitem[Lovett et~al., 2009b]{lovett2009solving}
Lovett, A., Tomai, E., Forbus, K., and Usher, J. (2009b).
\newblock Solving geometric analogy problems through two-stage analogical
  mapping.
\newblock {\em Cognitive science}, 33(7):1192--1231.

\bibitem[Markman and Gentner, 1993]{markman1993structural}
Markman, A.~B. and Gentner, D. (1993).
\newblock Structural alignment during similarity comparisons.
\newblock {\em Cognitive psychology}, 25(4):431--467.

\bibitem[McGrew, 2005]{mcgrew2005cattell}
McGrew, K.~S. (2005).
\newblock The cattell-horn-carroll theory of cognitive abilities: Past,
  present, and future.
\newblock In Flanagan, D.~P. and Harrison, P.~L., editors, {\em Contemporary
  Intellectual Assessment: Theories, Tests, and Issues}, pages 136--181. The
  Guilford Press.

\bibitem[Mensa et~al., 2018]{mensa2018grasping}
Mensa, E., Porporato, A., and Radicioni, D.~P. (2018).
\newblock Grasping metaphors: Lexical semantics in metaphor analysis.
\newblock In {\em European Semantic Web Conference}, pages 192--195. Springer.

\bibitem[Milkowski, 2013]{milkowski2013explaining}
Milkowski, M. (2013).
\newblock {\em Explaining the computational mind}.
\newblock Mit Press.

\bibitem[Mi{\l}kowski, 2026]{milkowski2026confirmation}
Mi{\l}kowski, M. (2026).
\newblock Confirmation and explanation in neuroscience: Reassessing the
  relationship between functional and mechanistic approaches.
\newblock In {\em Neurocognitive Foundations of Mind}, pages 38--58. Routledge.

\bibitem[Mostafazadeh et~al., 2016]{mostafazadeh2016corpus}
Mostafazadeh, N., Chambers, N., He, X., Parikh, D., Batra, D., Vanderwende, L.,
  Kohli, P., and Allen, J. (2016).
\newblock A corpus and cloze evaluation for deeper understanding of commonsense
  stories.
\newblock In {\em Proceedings of the 2016 Conference of the North American
  Chapter of the Association for Computational Linguistics: Human Language
  Technologies}, pages 839--849.

\bibitem[Musker et~al., 2024]{musker2024semantic}
Musker, S., Duchnowski, A., Milli{\`e}re, R., and Pavlick, E. (2024).
\newblock Semantic structure-mapping in llm and human analogical reasoning.
\newblock {\em arXiv preprint arXiv:2406.13803}.

\bibitem[Musker et~al., 2025]{musker2025llmsmodelsanalogicalreasoning}
Musker, S., Duchnowski, A., Millière, R., and Pavlick, E. (2025).
\newblock Llms as models for analogical reasoning.

\bibitem[Newell, 1994]{newell1994unified}
Newell, A. (1994).
\newblock {\em Unified theories of cognition}.
\newblock Harvard University Press.

\bibitem[OpenAI et~al., 2024]{openai2024gpt4technicalreport}
OpenAI, J.~A. et~al. (2024).
\newblock Gpt-4 technical report.

\bibitem[Otero, 2017]{otero2017brief}
Otero, T.~M. (2017).
\newblock Brief review of fluid reasoning: Conceptualization, neurobasis, and
  applications.
\newblock {\em Applied Neuropsychology: Child}, 6(3):204--211.

\bibitem[Paperno et~al., 2016]{paperno2016}
Paperno et~al. (2016).
\newblock The lambada dataset: Word prediction requiring a broad discourse
  context.

\bibitem[Pitt and Myung, 2002]{pittmyung2002}
Pitt, M.~A. and Myung, I.~J. (2002).
\newblock When a good fit can be bad.
\newblock {\em Trends in Cognitive Sciences}, 6(10):421--425.

\bibitem[Putnam, 1960]{putnam1960minds}
Putnam, H. (1960).
\newblock Minds and machines.
\newblock In Hook, S., editor, {\em Dimensions of Mind: A Symposium}, pages
  138--164. New York University Press, New York.

\bibitem[Quattrociocchi et~al., 2025]{quattrociocchi2025epistemological}
Quattrociocchi, W., Capraro, V., and Perc, M. (2025).
\newblock Epistemological fault lines between human and artificial
  intelligence.
\newblock {\em arXiv preprint arXiv:2512.19466}.

\bibitem[Raven, 1938]{raven1938}
Raven, J.~C. (1938).
\newblock {\em Progressive Matrices: A Perceptual Test of Intelligence}.
\newblock Lewis Raven, London.

\bibitem[Raven et~al., 1992]{raven1992manual}
Raven, J.~C., Court, J.~H., and Raven, J. (1992).
\newblock {\em Manual for Raven’s Progressive Matrices and Vocabulary
  Scales}.
\newblock Oxford Psychologists Press, Oxford.

\bibitem[Rosenblueth and Wiener, 1945]{rosenblueth1945role}
Rosenblueth, A. and Wiener, N. (1945).
\newblock The role of models in science.
\newblock {\em Philosophy of science}, 12(4):316--321.

\bibitem[Russell and Norvig, 2012]{russell1995artificial}
Russell, S.~J. and Norvig, P. (2012).
\newblock {\em Artificial intelligence: A modern approach;[the intelligent
  agent book]}.
\newblock Prentice hall.

\bibitem[Salthouse, 1993]{salthouse1993influence}
Salthouse, T.~A. (1993).
\newblock Influence of working memory on adult age differences in matrix
  reasoning.
\newblock {\em British Journal of Psychology}, 84(2):171--199.

\bibitem[Sternberg and Nigro, 1980]{sternberg1980developmental}
Sternberg, R.~J. and Nigro, G. (1980).
\newblock Developmental patterns in the solution of verbal analogies.
\newblock {\em Child Development}, pages 27--38.

\bibitem[Stevenson et~al., 2023]{stevenson2023large}
Stevenson, C.~E., ter Veen, M., Choenni, R., van~der Maas, H.~L., and Shutova,
  E. (2023).
\newblock Do large language models solve verbal analogies like children do?
\newblock {\em arXiv preprint arXiv:2310.20384}.

\bibitem[Team, 2024]{geminiteam2024geminifamilyhighlycapable}
Team, G. (2024).
\newblock Gemini: A family of highly capable multimodal models.

\bibitem[Turney et~al., 2003]{turney2003combining}
Turney, P.~D., Littman, M.~L., Bigham, J., and Shnayder, V. (2003).
\newblock Combining independent modules to solve multiple-choice synonym and
  analogy problems.
\newblock {\em arXiv preprint cs/0309035}.

\bibitem[Tversky, 1977]{tversky1977features}
Tversky, A. (1977).
\newblock Features of similarity.
\newblock {\em Psychological review}, 84(4):327.

\bibitem[Wachowiak and Gromann, 2023]{wachowiak-gromann-2023-gpt}
Wachowiak, L. and Gromann, D. (2023).
\newblock Does {GPT}-3 grasp metaphors? identifying metaphor mappings with
  generative language models.
\newblock In Rogers, A., Boyd-Graber, J., and Okazaki, N., editors, {\em
  Proceedings of the 61st Annual Meeting of the Association for Computational
  Linguistics (Volume 1: Long Papers)}, pages 1018--1032, Toronto, Canada.
  Association for Computational Linguistics.

\bibitem[Wang et~al., 2019]{Glue}
Wang, A. et~al. (2019).
\newblock Glue: A multi-task benchmark and analysis platform for natural
  language understanding.

\bibitem[Wang et~al., 2020]{superGlue}
Wang, A. et~al. (2020).
\newblock Superglue: A stickier benchmark for general-purpose language
  understanding systems.

\bibitem[Webb et~al., 2023]{webb2023emergent}
Webb, T., Holyoak, K.~J., and Lu, H. (2023).
\newblock Emergent analogical reasoning in large language models.
\newblock {\em Nature Human Behaviour}, 7(9):1526--1541.

\bibitem[Wilson, 2002]{wilson2002six}
Wilson, M. (2002).
\newblock Six views of embodied cognition.
\newblock {\em Psychonomic Bulletin \& Review}, 9(4):625--636.

\bibitem[Xu et~al., 2024]{xu2024lvlm}
Xu, P., Shao, W., Zhang, K., Gao, P., Liu, S., Lei, M., Meng, F., Huang, S.,
  Qiao, Y., and Luo, P. (2024).
\newblock Lvlm-ehub: A comprehensive evaluation benchmark for large
  vision-language models.
\newblock {\em IEEE Transactions on Pattern Analysis and Machine Intelligence}.

\bibitem[Ye et~al., 2024]{ye2024analobench}
Ye, X., Wang, A., Choi, J., Lu, Y., Sharma, S., Shen, L., Tiyyala, V., Andrews,
  N., and Khashabi, D. (2024).
\newblock Analobench: Benchmarking the identification of abstract and
  long-context analogies.
\newblock {\em arXiv preprint arXiv:2402.12370}.

\bibitem[Yin et~al., 2024]{yin2024lamm}
Yin, Z., Wang, J., Cao, J., Shi, Z., Liu, D., Li, M., Huang, X., Wang, Z.,
  Sheng, L., Bai, L., et~al. (2024).
\newblock Lamm: Language-assisted multi-modal instruction-tuning dataset,
  framework, and benchmark.
\newblock {\em Advances in Neural Information Processing Systems}, 36.

\bibitem[Yuan et~al., 2023]{yuan2023beneath}
Yuan, S., Chen, J., Ge, X., Xiao, Y., and Yang, D. (2023).
\newblock Beneath surface similarity: Large language models make reasonable
  scientific analogies after structure abduction.
\newblock {\em arXiv preprint arXiv:2305.12660}.

\bibitem[Zellers et~al., 2019]{hellaswag}
Zellers, R., Holtzman, A., Bisk, Y., Farhadi, A., and Choi, Y. (2019).
\newblock Hellaswag: Can a machine really finish your sentence?
\newblock {\em arXiv preprint arXiv:1905.07830}.

\bibitem[Zhang and Wang, 2024]{RPM-LLMs}
Zhang, C. and Wang, L. (2024).
\newblock Evaluating abstract reasoning and problem-solving abilities of large
  language models using raven's progressive matrices.

\bibitem[Zhong et~al., 2024]{zhong2024achieving}
Zhong, Q., Wang, K., Xu, Z., Liu, J., Ding, L., Du, B., and Tao, D. (2024).
\newblock Achieving> 97\% on gsm8k: Deeply understanding the problems makes
  llms perfect reasoners.
\newblock {\em arXiv e-prints}, pages arXiv--2404.

\end{thebibliography}

\end{document}